\documentclass{article}

\usepackage{microtype}
\usepackage{graphicx}
\usepackage{subfigure}
\usepackage{booktabs} 

\usepackage{hyperref}

\usepackage[accepted]{icml2024}

\usepackage{amsmath}
\usepackage{amssymb}
\usepackage{mathtools}
\usepackage{amsthm}
\usepackage{graphicx}
\usepackage{amssymb}
\usepackage{multirow}
\usepackage{pifont}
\usepackage{bbding}
\usepackage{amsmath}
\usepackage{color}
\usepackage{wasysym}
\usepackage{colortbl}  
\usepackage{xcolor}
\usepackage{array}  

\usepackage[capitalize,noabbrev]{cleveref}

\theoremstyle{plain}
\newtheorem{theorem}{Theorem}[section]
\newtheorem{proposition}[theorem]{Proposition}

\theoremstyle{definition}

\theoremstyle{remark}

\usepackage[textsize=tiny]{todonotes}

\icmltitlerunning{Pedestrian Attribute Recognition as Label-balanced Multi-label Learning}

\begin{document}
\twocolumn[
\icmltitle{Pedestrian Attribute Recognition as Label-balanced Multi-label Learning}




\begin{icmlauthorlist}
\icmlauthor{Yibo Zhou}{bh}
\icmlauthor{Hai-Miao Hu}{bh,bhy}
\icmlauthor{Yirong Xiang}{mu}
\icmlauthor{Xiaokang Zhang}{bh}
\icmlauthor{Haotian Wu}{bh}
\end{icmlauthorlist}

\icmlaffiliation{bh}{State key laboratory of virtual reality technology and systems, Beihang University, China}
\icmlaffiliation{bhy}{Hangzhou Innovation Institute, Beihang University, China}
\icmlaffiliation{mu}{The University of Manchester, UK}

\icmlcorrespondingauthor{Hai-Miao Hu}{hu@buaa.edu.cn}

\icmlkeywords{Machine Learning, ICML}

\vskip 0.3in
]



\printAffiliationsAndNotice{} 

\begin{abstract}
Rooting in the scarcity of most attributes, realistic pedestrian attribute datasets exhibit unduly skewed data distribution, from which two types of model failures are delivered: (1) \emph{label imbalance}: model predictions lean greatly towards the side of majority labels; (2) \emph{semantics imbalance}: model is easily overfitted on the under-represented attributes due to their insufficient semantic diversity. To render perfect label balancing, we propose a novel framework that successfully decouples label-balanced data re-sampling from the curse of attributes co-occurrence, i.e., we equalize the sampling prior of an attribute while not biasing that of the co-occurred others. To diversify the attributes semantics and mitigate the feature noise, we propose a Bayesian feature augmentation method to introduce true in-distribution novelty. Handling both imbalances jointly, our work achieves best accuracy on various popular benchmarks, and importantly, with minimal computational budget.
\end{abstract}

\section{Introduction}
In visual tasks, human attribute is generally not a precisely defined concept, and can encompass a spectrum of disparate soft-biometrics that range from locatable body parts to comprehensive human descriptors \cite{wang2022pedestrian, liu2017hydraplus}. Thus, for the pedestrian attribute recognition (PAR), it is inviable to craft a universal framework that efficiently yields level performance among myriad attributes of distinct characteristics. Specifically, for accessary attribute like hat or boot, the task of PAR essentially mirrors weakly supervised object detection \cite{zhang2021weakly}, as the model should infer on minimal area as relevant as possible for a discriminative recognition \cite{DBLP:conf/aaai/JiaGHCH22}. While for the attribute of whole-body semantics like action or ages, any explicit mechanism to discard spatial information may result in insufficiency of information exchange, revealing that in this case, PAR is more akin to regular visual classification.

As a result, the broadness of `attribute' implies it a loose umbrella term, and motivates us not to approach PAR from the perspective of over-specialized architectures \cite{lu2023orientation,Jia_2021_ICCV}. Thus, we question that, \emph{is there a more pervasive problem existing in realistic pedestrian attribute tasks, by solving which the predictions on overall attributes are expected to be evenly boosted?} Equipped with this aspiration, we concisely distill PAR into a problem of multi-label classification under significant data imbalance. 

\begin{figure}[t]
\centering
\includegraphics[height=3.9cm,width=8.05cm]{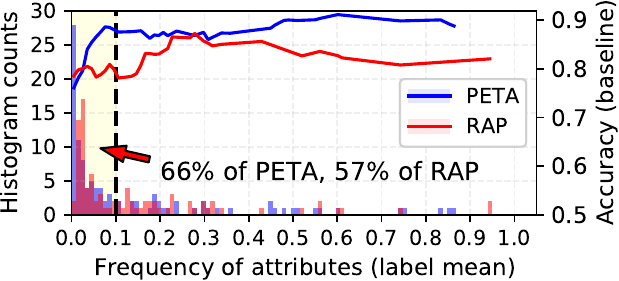}
\caption{The dominance of negative labels in PAR datasets, and mean accuracy as a function of the label mean. In PETA \cite{2014Pedestrian}, 66\% attributes occur with a frequency under 0.1, while that for RAP \cite{li2016richly} is 57\%. Also, label imbalance is the main performance bottleneck of contemporary PAR model, as it is significantly brittle to attributes with label mean $\le$ 0.1.}
\label{label_mean}
\end{figure}

This simplification makes sense as: (1) the ambiguity and variety within attributes require a much general PAR definition; (2) since most attributes occur with small empirical frequencies, PAR datasets are profoundly label-imbalanced. Worse, attribute label priors exhibit great unpredictability across various sceneries \cite{zhou2023solution}, making it an immense data selection bias that hardly generalizes; (3) previous work only partly alleviates label imbalance by experimentally setting different penalizing weights for labels \cite{wang2022pedestrian}, or just abstains the overly infrequent attributes from benchmarks to display decent overall results \cite{jia2021rethinking}. Thus, label imbalance is de facto the \emph{grey rhino} that fundamentally bottlenecks the performance of PAR (Figure \ref{label_mean}), and remains critically under-addressed. 

While data re-sampling (over-sample the images from under-represented label or under-sample the others) can facilitate unbiased label prior for long-tailed recognition \cite{zhang2021bag}, it is infeasible to be directly adopted into PAR owing to the intricate entanglement of attributes in images. In specific, given the limited patterns of label co-occurrence in dataset, repeating/dropping images to equalize the sampling distribution of one attribute will inevitably skew the balance of others \cite{guo2021long}. In contrast to segmenting attributes in pixel space for independent sampling, we demonstrate that such a gap can be absolutely bridged if the re-sampling space is shifted from images to latent features. Consequentially, we develop a novel training pipeline of feature re-sampled learning to offer immunity to this curse of label co-occurrences, and thus ensure true \textbf{label balance} for PAR. Both theoretical insights and experimental evidence suggest that our method is an ideal drop-in instantiation for the intractable label-balanced image re-sampling of PAR. 

However, since the features of under-represented attributes may not suffice to describe the intra-class expressivity, when they are over-repeatedly sampled for label balancing, severe overfitting can be posed. To palliate such incidental overfitting, we aim to enrich feature novelty to attain \textbf{semantics balance}. One principled solution for it is resorting to feature augmentation techniques \cite{devries2017dataset}, and a prevalent recipe in this topic is built with an implicit assumption that the intra-class translating direction is homogeneous across the feature space, and samples synthetic points from identical gaussian clouds centering at different features \cite{wang2019implicit, li2021metasaug}. 

Unfortunately, we unveil that no novel variety is introduced by these homogeneous methods as they can be essentially reformulated as large-margin optimizers \cite{liu2016large} with static margins. As a counter, we state the necessity of heterogeneous feature augmentation for genuine semantics diversification, and promote a Bayesian method for it. With our approach, feature of impoverished labels is augmented by non-trivial gradient-based stochasticity, in effect relieving the exacerbated overfitting. Also, we theoretically prove that our method is able to assuage the data noise from spurious feature during feature re-sampling.

Coping with both the label imbalance and semantics imbalance in a highly holistic manner, our method surpasses prior arts with considerable margins, and establishes state-of-the-art performance on various benchmarks. Albeit effective, our prescription is desirably lightweight as minimal extra parameters are entailed. Our contribution is three-fold:

\begin{itemize} 
\item  To our best knowledge, this is the first work that develops true label-balanced learning for multi-label tasks.

\item  We elaborate on the whys and wherefores of the pitfall of existing feature augmentation methods, and propose a Bayesian approach to create true novel features. 

\item By mitigating two types of imbalance, our lightweight framework scores best w.r.t. mean accuracy on realistic PAR benchmarks. Extensive ablation and robustness studies also validate a suite of merits of our proposal.

\end{itemize}

\section{Related Work}
\textbf{Pedestrian Attribute Recognition.} Basically, there are two common paradigms in PAR. First class of studies has delved into enhancing attributes localization to reduce the accuracy drop from predicting on extraneous area. Various attention mechanisms \cite{liu2018localization,DBLP:conf/aaai/JiaGHCH22,liu2017hydraplus}, attributes partition strategies \cite{fabbri2017generative,2017Learning} and body-attributes dependencies \cite{liu2018localization,lu2023orientation} were leveraged to better capture the spatial topological structure of attributes. Another active research stream regards attributes correlation as a concrete prior \cite{2022Label2Label,2020Correlation,wang2017discovering}, and attempts to exploit attributes interdependencies by graph models. However, both lines of work are questionable. \cite{2020Rethinking} showed that attribute positioning may not be the core performance bottleneck of PAR. Also, \cite{zhou2023solution} discovered that attributes co-occurrence is more like a mutable data selection bias that impairs the PAR performance. Such paradoxical results make us rethink, what is indeed a fundamental factor for PAR to scale well?

\textbf{Imbalance in Multi-label Tasks.} Limited by the label co-occurrences, existing multi-label methods ease the label imbalance mainly by loss re-weighting \cite{jia2021rethinking}, such as using the inverse of label-wise sample size in loss function to up-weight minority data \cite{xu2022adaptive}, or other alternative weighting functions \cite{li2015multi, tan2020relation}. Differently, this work achieves label-balanced re-sampling for multi-label recognition. Moreover, not only the numerical asymmetry of labels distribution, we also milden the twined semantics imbalance.

\section{Method}
\subsection{On the Label-balanced Re-sampling of PAR} \label{m11}
Formally, let $X$ be a distribution characterized by all of the pedestrian surveillance images. Some data points $\{ \boldsymbol x_{i} \}_{i=1}^{N}$ are sampled from $X$, jointly with their corresponding labels $\{ \boldsymbol y_{i} \}_{i=1}^{N}$ of certain attributes to form a dataset $D$, where $N$ denotes the dataset cardinality $|D|$, $\boldsymbol y_{i} \in \{0, 1\}^C$ and $C$ is the number of total annotated attributes. Each element in $\boldsymbol y_i$ serves as the 0/1 indicator of the occurrence of an attribute in $\boldsymbol x_i$. Practically, such a dataset $D$ is collected from $X$ with small empirical attribute frequencies. It results in that $\frac{N^k}{N}, \forall k = 1,2,...,C$, can be far from 0.5, where $N^k$ is the number of images in $D$ with attribute label $\boldsymbol y^{k}$ being 1. Consequentially, the separating hyperplane in the decision space will be heavily skewed to the label of relatively few number, from where poor PAR performance is delivered.

Label-balanced re-sampling is the most straightforward approach to facilitate recognition with such imbalanced labels.
\begin{center}
\fbox
{\shortstack[c]{
\textbf{Label-balanced Image Re-sampling (LIR)}: Adjust the\\
sampling function of images, to let the attributes images\\
fed into model perfectly balanced between binary labels.
}
}
\end{center}

LIR is achievable only if there exists $\{a_i\}_{i=1}^{N}$ satisfying 

\begin{equation}
\begin{aligned}
&\sum_{i=1}^{N} \boldsymbol y_i \cdot a_i + \sum_{i=1}^{N} (\boldsymbol y_i-\boldsymbol 1) \cdot a_i = \boldsymbol 0,\\
\text{  s.t.  }&\sum_{i=1}^{C} a_i = 1, \,\,\, a_i > 0, i = 1,2,...,N.
\label{condition}
 \end{aligned}
\end{equation}

Since patterns of attributes co-occurrence can be quite limited \cite{zhou2023solution}, gathering a dataset meeting Eq.\ref{condition} is difficult. It reveals that, re-adjusting the sampling function of a certain attribute to balance its label prior would yield another biased distribution for others. Also, as $a_i$ represents the probability of $\boldsymbol x_i$ to be sampled, it is expected that all $a_i > 0$ and have a similar value such that data points can be sampled with comparable odds, making an acceptable sampling function much impracticable to get. Essentially, such curse of label co-occurrence roots from that all attributes are entangled in input images, implying that for independent balancing of each attribute, LIR would be preconditioned on some challenging methods to precisely segment attributes in pixel space. Instead of attributes segmenting, we attempt to label-balanced re-sample attributes in a label-disentanglable feature space to unconditionally bridge this gap.

 
\subsection{Feature Re-sampled Decoupled Learning} 
For multi-class recognition, decoupling is one of the training schemes most successful on long-tailed datasets \cite{zhang2023deep}. Its two-stage workflow is streamlined as

\begin{center}
\fbox
{\shortstack[c]{
\textbf{Decoupled Learning (DL)}: \emph{Stage\#1}: Do vanilla training \\
with instance-balanced sampled images to learn a whole\\
model. \emph{Stage\#2}: The images are label-balanced sampled,\\ 
and only fine-tune the classifier with other modules fixed. 
}
}
\end{center}

Compared to label-balanced image re-sampling, DL renders better accuracy on long-tailed dataset, since it not only gives same neutral decision boundaries in classifier, but also produces more discriminative latent representations thanks to that the feature extractor in DL is not overfitting on the over-sampled images of minority classes \cite{kang2019decoupling}. 

Inspired by it, we conjecture that solving the impossibility of label-balanced sampling in the attributes-entangled pixel space might not be technically indispensable for true balanced PAR, as we actually do not need a label-balanced learned feature extractor. In other words, \emph{all we need is a label-balanced classifier.} Importantly, this concept remedies the curse of attributes co-occurrence of LIR for PAR, \emph{since unlike feature extractor, classifiers weight is not shared among attributes, meaning that the inferences of attributes are already structurally disentangled in the final classification step, and independent attributes re-sampling is thus viable for PAR classifier.} To this end, we devise the pipeline

\begin{center}
\fbox
{\shortstack[c]{
\textbf{Feature Re-sampled Decoupled Learning (FRDL)}:\\
\emph{Stage\#1}: Do vanilla training with instance-balanced data\\
 sampling to learn a whole model. \emph{Stage\#2}: Input image is\\
 still instance-balanced sampled and fed into fixed feature\\
extractor to produce representations. Differently, features\\
are saved in memory banks according to their labels, and\\
classifier is re-trained on label-balanced sampled features.
}
}
\end{center}

\begin{figure*}[t]
\centering
\includegraphics[height=3.85cm,width=16.12cm]{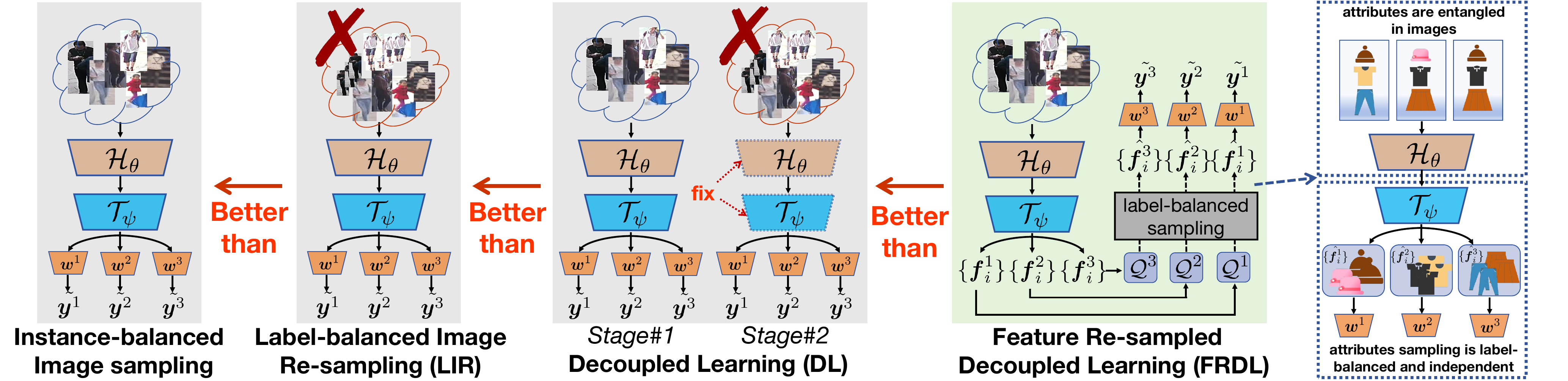}
\caption{Schematic presentation of the main idea of FRDL. Although DL can not be naively implemented for PAR due to the unsatisfiable label-balanced image re-sampling (Eq.\ref{condition}), its better form of FRDL is workable and thus acts as a better drop-in substitution of LIR.}
\label{frdl}
\end{figure*}

as an upper substitution of LIR. Concretely, we denote with $\boldsymbol f_i = \mathcal H_{\theta}(\boldsymbol x_i) \in \mathbb R^M$ the representation of $\boldsymbol x_i$, where $\mathcal H_{\theta}(\cdot)$ is a feature extractor parameterized by $\theta$. Sequentially, $\boldsymbol f_i$ is decomposed into $M$-dimensional attribute-specific features $\{\boldsymbol f_i^{k}\}_{k=1}^C = \mathcal T_{\psi}(\boldsymbol f_i)$ by a fully-connected layer $T_{\psi}(\cdot)$. Attribute posterior is finally estimated with a linear classifier function $\tilde{\boldsymbol y_i^k} = \boldsymbol w^k \boldsymbol f_i^k +b^k$, where $\boldsymbol w^k \in \mathbb R^{M}$ represents the classifier weight, and $b^k \in \mathbb R$ the bias, $\forall k = 1,2,...,C$. 

For \emph{Stage\#1}, we train whole model on the instance-balanced sampled images by plain binary cross-entropy (BCE) loss. When model converges, we feed the whole dataset $\{ \boldsymbol x_{i} \}_{i=1}^{N}$ into fixed $\mathcal H_{\theta}(\cdot)$ and $\mathcal T_{\psi}( \cdot)$, and collect the output representations $\{(\boldsymbol f_{i}^1,\boldsymbol f_{i}^2,...,\boldsymbol f_{i}^C)\}_{i=1}^{N}$ into $C$ pairs of attribute-specific feature banks $\{(Q_0^k, Q_1^k)\}_{k=1}^C$. Specifically, $Q_0^k$ and $Q_1^k$ save all $\boldsymbol f_i^k$ with label $\boldsymbol y_i^k $ being 0 and 1, respectively. Finally, the \emph{Stage\#2} of FRDL draws between $Q_0^k$ vs. $Q_1^k$ with an equal probability, and a feature from the selected bank is uniformly sampled with replacement to form a label-balanced training batch, atop which $(\boldsymbol w^k, b^k)$ is fine-tuned. 

Seemingly, FRDL and DL make no difference in multi-class tasks. However, in the context of PAR, FRDL is non-trivial as: (1) it unconditionally achieves label-balanced classifier, by transferring the unsatisfiable label-balanced image re-sampling in the \emph{Stage\#2} of DL to a tractable label-balanced feature re-sampling; (2) even if Eq.\ref{condition} is satisfied for DL, the over-sampled images to balance an attribute will be uncalled-for repeated in the classifiers learning of other attributes, propagating the overfitting issue coupled with balanced re-sampling of one attribute to all attributes. Differently, as the attributes inferences are already disentangled in classifier, FRDL enables not only label-balanced, but also independent classifier learning for each attribute, and thus performs better than DL. Main concept of FRDL is illustrated in Figure \ref{frdl}.

\subsection{Pitfall of Homogeneous Feature Augmentation} 

To obviate overfitting aggravated by the over-sampled features in FRDL, an intuitive solution is to diversify the limited statistics of minority attributes by strong data augmentation \cite{chawla2002smote}. However, most image augmentation techniques can potentially obliterate the delicate signatures of small attributes within the pixel space, thus leading to subpar performance (see Appendix \ref{D3}). Hence, we resort to augment data in latent space. Postulating that certain directions in feature space are aligned with intra-class semantics variation, ISDA \cite{wang2019implicit} and its follow-ups translate the features linearly in some latent directions to augment additional representations. For PAR, they can be expressed as a feature re-sampling process of $\tilde{\boldsymbol f_i^k} \sim \mathcal N(\boldsymbol f_i^k, \lambda^k \boldsymbol \Sigma^k )$, and just differ by the specific choice of $\{\boldsymbol \Sigma^k\}_{k=1}^N$.

Since the translating directions at different $\boldsymbol f_i^k$ are sampled from a same prior $\mathcal N(0, \lambda^k \boldsymbol \Sigma^k )$, they are actually presumed, by ISDA, homogeneous across the whole latent space. However, on one hand latent direction of intra-class variation is not as homogeneous as consistent gaussian clouds, since in practice features are distributed heterogeneously \cite{wan2018rethinking}. On the other hand, to explore all directions in $\mathcal N(0, \lambda^k \boldsymbol \Sigma^k )$, one should minimize the expectation of the BCE loss of PAR, under all possible augmented features, as \vspace{-3.0ex} 

\begin{equation}
\begin{aligned}
&\mathbb E_{\tilde{\boldsymbol f_i^k}}[\frac{1}{N}\sum_{i=1}^N\sum_{k=1}^C \log(1+e^{-\mathbb{I}(\boldsymbol y^k_i)\cdot(\boldsymbol w^k \tilde{\boldsymbol f_i^k}+b^k)})] \\
&\le \frac{1}{N}\sum_{i=1}^N\sum_{k=1}^C \log\mathbb E_{\tilde{\boldsymbol f_i^k}}[1 + e^{-\mathbb{I}(\boldsymbol y^k_i)\cdot(\boldsymbol w^k \tilde{\boldsymbol f_i^k}+b^k)}] \\
&= \rlap{$\underbrace{\phantom{\frac{1}{N}\sum_{i=1}^N\sum_{k=1}^C \log(1 + e^{-\mathbb{I}(\boldsymbol y^k_i)\cdot(\boldsymbol w^k \boldsymbol f_i^k + b^k)}}}_{\text{binary cross-entropy on $\boldsymbol f_i^k$}}$}\frac{1}{N}\sum_{i=1}^N\sum_{k=1}^C \log(1 + e^{-\mathbb{I}(\boldsymbol y^k_i)\cdot(\boldsymbol w^k \boldsymbol f_i^k + b^k) + \frac{1}{2}\boldsymbol w^{k\top} \lambda^k \boldsymbol \Sigma^k \boldsymbol w^k}).
\label{constant}
\end{aligned}
\end{equation}

In Eq.\ref{constant}, $\mathbb{I}(\boldsymbol y_i^k) = 1$ if $\boldsymbol y_i^k = 1$ and $\mathbb{I}(\boldsymbol y_i^k) = -1$ if $\boldsymbol y_i^k = 0$. The inequality follows from the Jensen inequality and the final step is obtained by the moment-generating function for the gaussian variable $\tilde{\boldsymbol f_i^k}$. It reveals that Eq.\ref{constant}, a closed-form upper bound of the homogeneous feature augmentation loss, is in essence a vanilla BCE loss with fixed inter-label margins since $\{\frac{1}{2}\boldsymbol w^{k\top} \lambda^k \boldsymbol \Sigma^k \boldsymbol w^k\}_{k=1}^C$ are just constants. Thus, \emph{homogeneous methods are endogenously large-margin optimizers}, and carefully tuning $\{\lambda^k\}_{k=1}^C$ like their original paper is intrinsically enumerating the priori-unknown best inter-label margin and will finally smooth out any difference in the specific choices of $\{\boldsymbol \Sigma^k\}_{k=1}^N$. As a result, we argue that no novel diversity regarding distribution exploration can be inherently introduced by homogeneous methods.

\subsection{Gradient-oriented Augment Translating} 

We are now in a position to overcome above issue. A desirable translating direction to augment features should comprise: (1) \textbf{in-distribution}, the augmented features still reside in the latent domain of same attribute identity; (2) \textbf{meaningful}, the translating directions co-linear with attribute semantics shifting, instead of some random noise; (3) \textbf{heterogeneous}, the translating direction of each feature is computed from its own neighborhood of the distribution. Hence, for any feature point $\boldsymbol f_i^k$ within a trained model, we translate it along its local gradient to augment new feature

\begin{equation}
\begin{aligned}
\tilde{\boldsymbol f_i^k} = \boldsymbol f_i^k - \eta \nabla_{\boldsymbol f^k = \boldsymbol f_i^k} |\mathcal L_{cls}(\boldsymbol f^k) - \mathbb E_{\boldsymbol f^k}[\mathcal L_{cls}(\boldsymbol f^k)]|,
\label{augment}
\end{aligned}
\end{equation}

where $\mathcal L_{cls}(\cdot)$ computes the BCE loss of $\boldsymbol f$, and $\eta$ is a positive step size. During this process, the classifier utilized for the gradient computation is well-trained and remains fixed. Conversely, a fresh classifier is independently trained from scratch with $\tilde{\boldsymbol f_i^k}$, and finally takes over for the test-time classification. The rationales behind applying Eq.\ref{augment} for feature augmentation are: (1) the translating is high-density oriented as it always points to the distribution centroid $\mathbb E_{{\boldsymbol f^k}}[\mathcal L_{cls}({\boldsymbol f^k})]$. Therefore, the over-confident features (small loss) would be pulled back to be less-confident, while the noisy features (large loss) would be relaxed into high-density zone. Consequentially, no outliers are created, leading to \textbf{in-distribution}; (2) the feature is transferred in the direction of loss gradient, which is most relevant to the attribute informativeness across the entire space. It enables that, instead of a quasi replication, the augmented feature is novel w.r.t. its initial representation in term of the embedded attributes semantics, i.e., the translating is \textbf{meaningful}; (3) with subsequent non-linear classifier, the gradient varies among different feature points, making Eq.\ref{augment} form a \textbf{heterogeneous} sampling field of translating directions. 

Practically, the proposed Gradient-Oriented Augment Translating (GOAT) in Eq.\ref{augment} can be seamlessly implemented without further efforts. In specific, if we optimize the feature extractor $\mathcal T_{\psi_t}(\mathcal H_{\theta_t}(\cdot))$ at training step $t$ by gradient descend w.r.t. a succinct loss $\mathcal L_{goat}$ of

\begin{equation}
\begin{aligned}
& \,\,\,\,\, \frac{1}{N}\sum_{i=1}^N\sum_{k=1}^C |\mathcal L_{cls}(\boldsymbol f_{i,t}^k) - \mu^k_t|,\\
&s.t.\,\,\,\,\, \{\boldsymbol f_{i,t}^k\}_{k=1}^C = \mathcal T_{\psi_t}(\mathcal H_{\theta_t}( \boldsymbol x_i)),
\label{augloss}
\end{aligned}
\end{equation}

where $\mu^k_{t}$ is $\mathbb E_{\boldsymbol f^k_t}[\mathcal L_{cls}(\boldsymbol f^k_{t})]$, sequentially, $\tilde{\boldsymbol f_{i,t}^k}$ that translated from $\boldsymbol f_{i,t}^k$ by Eq.\ref{augment} would be identical to $\boldsymbol f_{i,t+1}^k$ generated by $\mathcal T_{\psi_{t+1}}(\mathcal H_{\theta_{t+1}}(\cdot))$. The reason is that, to minimize Eq.\ref{augloss}, the feature extractor would be updated to translate $\boldsymbol f_{i,t}^k$ along the same direction of $-\nabla_{\boldsymbol f_t^k = \boldsymbol f_{i,t}^k} |\mathcal L_{cls}(\boldsymbol f_t^k) - \mathbb E_{\boldsymbol f_t^k}[\mathcal L_{cls}(\boldsymbol f_t^k)]|$ in Eq.\ref{augment}. Importantly, \emph{it reveals the inherent equivalence between gradient-oriented feature augmentation and the feature extractor gradient-descending.} Thus, to incorporate additional stochasticity, we optimize ($\theta_{t_0}, \psi_{t_0}$), which is the optimum feature extractor pre-trained on $D$, w.r.t. Eq.\ref{augloss} and treat the features collected along a short stochastic gradient descent (SGD) trajectory of $\{\theta_{t_0+s}, \psi_{t_0+s}\}_{s=0}^{T}$ as representations aptly augmented from the features at $t_0$, where $T \ge 1$. As such, GOAT approximates Bayesian feature sampling, as we can use $(\theta, \psi)$ at different steps to produce the probabilistic representations of a same input data. In this regard, GOAT essentially constructs a high-density-oriented \emph{heterogeneous Bayesian sampling cloud} around $\boldsymbol f_i$, which is in contrast to the homogeneous sampling cloud of prior arts. Notably, throughout the entire process, the likely feature distortion towards out-of-distribution is mitigated, since the classifier for gradient computation is fixed, resulting in that the subsequent $\{\theta_{t_0+s}, \psi_{t_0+s}\}_{s=1}^{T}$ would evolve within the vicinity of the initial classifier solution. Also, we set $T$ as a small number, and reload the model with $(\theta_{t_0}, \psi_{t_0})$ when the SGD trajectory reaches $T$ (larger $T$ produces stochasticity beyond Eq.\ref{augment}, but with more risk of off-distribution).

\begin{algorithm}[tb]
\caption{Pseudo-code of the GOAT-enhanced FRDL}
\label{alg}
\hspace*{0.02in} {\bf Input:} 
Training set $D$; ending step $T_1$, $T_2$ and $T$; initialized modules $\mathcal H_{\theta_0}( \cdot)$ and $\mathcal T_{\psi_0}( \cdot)$;  initialized classifiers $\mathcal G^{cls}_{W_0}( \cdot)$ and  $\mathcal G^{ft}_{W_0}(\cdot)$; empty feature banks $\{(Q_0^a, Q_1^a)\}_{a=1}^C$

\hspace*{0.02in} {\bf Output:} 
Instanced-balanced $\mathcal H_{\theta_{T_1}}( \cdot), \mathcal T_{\psi_{T_1}}( \cdot)$; Label-balanced $\mathcal G^{ft}_{W_{T_2}}(\cdot)$ \,\,\,\,\,\textcolor{gray}{\# final model is $\mathcal G^{ft}_{W_{T_2}}(\mathcal T_{\psi_{T_1}}(\mathcal H_{\theta_{T_1}}(\cdot)))$}

\begin{algorithmic}[1]
\STATE \textbf{for\,\,}$j=1$ {\bfseries to} $T_1$ \textbf{do}  \,\,\,\,\,\,\,\,\,\,\,\,\,\,\,\,\,\,\,\,\,\,\,\,\,\,\,\,\,\,\,\,\,\,\,\,\,\,\textcolor{gray}{\#\emph{Stage\#1} for FRDL}
\STATE \,\,\,\,\,\,Instance-balanced draw a batch $B=\{(\boldsymbol{x}_i, \boldsymbol y_i)\}_{i=1}^{|B|}$
\STATE \,\,\,\,\,\,Calculate loss \,\,\,\, \textcolor{gray}{\# can be any alternative of PAR loss}\\ 
\,\,\,\,\,\,$\mathcal{L}_{B} = \frac{1}{|B|}\sum_{i=1}^{|B|}L_{bce}(\mathcal G^{cls}_{W_{j-1}}(\mathcal T_{\psi_{j-1}}(\mathcal H_{\theta_{j-1}}(\boldsymbol{x}_i))),\boldsymbol y_i)$
\STATE \,\,\,\,\,\,Update: $W_{j}\leftarrow W_{j-1}-\alpha\nabla_{W}\mathcal{L}_{B}$; \textcolor{gray}{\# $\alpha$: learning rate}\\
\,\,\,\,\,\,$\theta_{j}\leftarrow\theta_{j-1}-\alpha\nabla_{\theta}\mathcal{L}_{B}$; $\psi_{j}\leftarrow\psi_{j-1}-\alpha\nabla_{\psi}\mathcal{L}_{B}$
\STATE Compute the attribute-wise loss centroid {$\mu^b$} in Eq.\ref{augloss} by $\mathcal G^{cls}_{W_{T_1}}(\cdot), \mathcal H_{\theta_{T_1}}( \cdot)$ and $\mathcal T_{\psi_{T_1}}( \cdot)$ on $D$, $\forall b = 1,2,...,C$
\STATE \textbf{for\,\,}$j=T_1$ {\bfseries to} $T_2$ \textbf{do} \,\,\,\,\,\,\,\,\,\,\,\,\,\,\textcolor{gray}{\#\emph{Stage\#2} for FRDL + GOAT}
\STATE \,\,\,\,\,\,Instance-balanced draw a batch $B=\{(\boldsymbol{x}_i, \boldsymbol y_i)\}_{i=1}^{|B|}$
\STATE \,\,\,\,\,\,\textbf{if} $(j \,\%\, T = 0$ \textbf{or} $j = T_1): k = T_1$ \textbf{ else}: $k=k+1$
\STATE \,\,\,\,\,\,Calculate $\mathcal L_{goat}$ on $B$ w.r.t. $\mathcal G^{cls}_{W_{T_1}}(\mathcal T_{\psi_{k}}(\mathcal H_{\theta_{k}}(\cdot)))$ and\\
\,\,\,\,\,\,$\{\mu^b\}_{b=1}^C$ by Eq.\ref{augloss}
\STATE \,\,\,\,\,\,Update: \,\,\,\,\,\,\,\,\,\,\,\,\,\,\textcolor{gray}{\# optimize $\theta \& \psi$ around $W_{T_1}$ solution}\\
\,\,\,\,\,\,$\theta_{k+1}\leftarrow\theta_{k}-\alpha\nabla_{\theta}\mathcal{L}_{goat}$; $\psi_{k+1}\leftarrow\psi_{k}-\alpha\nabla_{\psi}\mathcal{L}_{goat}$
\STATE \,\,\,\,\,\,Save the produced features into $\{(Q_0^a, Q_1^a)\}_{a=1}^C$
\STATE \,\,\,\,\,\,\textbf{for\,\,}$l=1$ {\bfseries to} $C$ \textbf{do} \,\,\,\,\,\,\,\,\,\,\,\,\,\textcolor{gray}{\#label-balanced train $\mathcal G^{ft}(\cdot)$}
\STATE \,\,\,\,\,\,\,\,\,\,\,\,Form a batch $\hat{B}=\{(\tilde{\boldsymbol{f}^l_i}, \boldsymbol y^l_i)\}_{i=1}^{|\hat{B}|}$ label-balanced \\
\,\,\,\,\,\,\,\,\,\,\,\,drawn from $(Q_0^l, Q_1^l)$ with replacement
\STATE \,\,\,\,\,\,\,\,\,\,\,\,Calculate loss \,\,\,\,\,\,\,\,\,\textcolor{gray}{\# $W^l$ denotes $l$-th column of $W$}\\ 
\,\,\,\,\,\,\,\,\,\,\,\,$\mathcal{L}_{\hat{B}} = \frac{1}{|\hat{B}|}\sum_{i=1}^{|\hat{B}|}|\mathcal L_{cls}^{W_j^l}(\tilde{\boldsymbol{f}^l_i}) - \mu^l |$
\STATE \,\,\,\,\,\,\,\,\,\,\,\,Update: $W^l_{j+1}\leftarrow W^l_{j}-\alpha\nabla_{W^l}\mathcal{L}_{\hat{B}}$
\end{algorithmic}
\end{algorithm}

\begin{table*}
\caption{Comparisons with the state-of-the-art methods. We refer the strong baseline results on PA100k and PETA from \cite{Specker2022UPARUP}, and the baseline results on RAPv1 are from our experiments. All values are percentages, and the best are highlighted in \textbf{bold} fonts.} 
\label{benchmark}
\large
\centering
\scalebox{0.75}{
        \begin{tabular}{cccc|ccc|cc|cc|cc}
        \toprule
	\multicolumn{4}{c|}{\multirow{2}{*}{Method}}&\multicolumn{3}{c|}{\multirow{2}{*}{Network}}&\multicolumn{2}{c|}{PA100k}&\multicolumn{2}{c|}{RAPv1}&\multicolumn{2}{c}{PETA}\\	
	\cline{8-13}
	\multicolumn{4}{c|}{}&\multicolumn{3}{c|}{}&\multicolumn{1}{c}{mA}&\multicolumn{1}{c|}{F1}&\multicolumn{1}{c}{mA}&\multicolumn{1}{c|}{F1}&\multicolumn{1}{c}{mA}&\multicolumn{1}{c}{F1} \\
 	\hline
	\hline
	\multicolumn{4}{c|}{\multirow{3}{*}{Baseline \cite{Specker2022UPARUP}}}&\multicolumn{3}{c|}{\emph{ResNet-50}}&\multicolumn{1}{c}{81.6-}&\multicolumn{1}{c|}{88.1-}&\multicolumn{1}{c}{80.18}&\multicolumn{1}{c|}{79.32}&\multicolumn{1}{c}{84.0-}&\multicolumn{1}{c}{86.3-}\\
	\multicolumn{4}{c|}{}&\multicolumn{3}{c|}{\emph{ConvNeXt-base}}&\multicolumn{1}{c}{82.2-}&\multicolumn{1}{c|}{88.5-}&\multicolumn{1}{c}{80.61}&\multicolumn{1}{c|}{81.76}&\multicolumn{1}{c}{86.1-}&\multicolumn{1}{c}{88.1-}\\
	\multicolumn{4}{c|}{}&\multicolumn{3}{c|}{\emph{Swin Transformer}}&\multicolumn{1}{c}{83.2-}&\multicolumn{1}{c|}{88.5-}&\multicolumn{1}{c}{82.12}&\multicolumn{1}{c|}{\textbf{82.30}}&\multicolumn{1}{c}{86.6-}&\multicolumn{1}{c}{87.7-}\\
	\hline
	\multicolumn{4}{c|}{DAFL \cite{DBLP:conf/aaai/JiaGHCH22}}&\multicolumn{3}{c|}{\emph{ResNet-50}}&\multicolumn{1}{c}{83.54}&\multicolumn{1}{c|}{\textbf{88.90}}&\multicolumn{1}{c}{83.72}&\multicolumn{1}{c|}{80.29}&\multicolumn{1}{c}{87.07}&\multicolumn{1}{c}{86.40}\\
	\multicolumn{4}{c|}{DRFormer \cite{tang2022drformer}}&\multicolumn{3}{c|}{\emph{ViT-B/16}}&\multicolumn{1}{c}{82.47}&\multicolumn{1}{c|}{88.04}&\multicolumn{1}{c}{81.81}&\multicolumn{1}{c|}{81.42}&\multicolumn{1}{c}{\textbf{89.96}}&\multicolumn{1}{c}{88.30}\\
	\multicolumn{4}{c|}{IAA-Caps \cite{wu2022inter}}&\multicolumn{3}{c|}{\emph{OSNet}}&\multicolumn{1}{c}{81.94}&\multicolumn{1}{c|}{87.80}&\multicolumn{1}{c}{81.72}&\multicolumn{1}{c|}{80.37}&\multicolumn{1}{c}{85.27}&\multicolumn{1}{c}{85.64}\\
	\multicolumn{4}{c|}{EALC \cite{weng2023exploring}}&\multicolumn{3}{c|}{\emph{EfficientNet-B4}}&\multicolumn{1}{c}{81.45}&\multicolumn{1}{c|}{88.14}&\multicolumn{1}{c}{83.26}&\multicolumn{1}{c|}{81.67}&\multicolumn{1}{c}{86.84}&\multicolumn{1}{c}{88.40}\\
	\multicolumn{4}{c|}{DFDT \cite{zheng2023diverse}}&\multicolumn{3}{c|}{\emph{Swin Transformer}}&\multicolumn{1}{c}{83.63}&\multicolumn{1}{c|}{88.74}&\multicolumn{1}{c}{82.34}&\multicolumn{1}{c|}{82.15}&\multicolumn{1}{c}{87.44}&\multicolumn{1}{c}{88.19}\\
	\multicolumn{4}{c|}{FEMDAR \cite{cao2023novel}}&\multicolumn{3}{c|}{\emph{ResNet-50}}&\multicolumn{1}{c}{81.02}&\multicolumn{1}{c|}{87.32}&\multicolumn{1}{c}{79.71}&\multicolumn{1}{c|}{78.76}&\multicolumn{1}{c}{84.73}&\multicolumn{1}{c}{85.90}\\
	\multicolumn{4}{c|}{VAL-PAT \cite{bao2023learning}}&\multicolumn{3}{c|}{\emph{\normalsize{ResNet-50\&Transformer}}}&\multicolumn{1}{c}{82.3-}&\multicolumn{1}{c|}{88.5-}&\multicolumn{1}{c}{80.8-}&\multicolumn{1}{c|}{81.0-}&\multicolumn{1}{c}{83.1-}&\multicolumn{1}{c}{84.4-}\\
	\multicolumn{4}{c|}{PARFormer-B \cite{fan2023parformer}}&\multicolumn{3}{c|}{\emph{Swin Transformer}}&\multicolumn{1}{c}{83.95}&\multicolumn{1}{c|}{87.69}&\multicolumn{1}{c}{83.84}&\multicolumn{1}{c|}{81.16}&\multicolumn{1}{c}{88.65}&\multicolumn{1}{c}{88.66} \\
	\hline
	\rowcolor{lightgray!30} \multicolumn{4}{c|}{\textbf{FRDL} + \textbf{GOAT}}&\multicolumn{3}{c|}{\emph{ConvNeXt-base}}&\multicolumn{1}{c}{\textbf{89.44}}&\multicolumn{1}{c|}{88.05}&\multicolumn{1}{c}{\textbf{87.72}}&\multicolumn{1}{c|}{79.16}&\multicolumn{1}{c}{88.59}&\multicolumn{1}{c}{\textbf{89.03}}\\
	 \bottomrule
\end{tabular}}
\end{table*}

Eq.\ref{augloss} also can be reasoned and adopted within the setting of feature de-noising. Practically, no matter how well-trained a feature extractor is, it inevitably encounters failures in pinpointing certain attributes from hard-case images, deriving information from the background as discriminative attribute representations and mistakenly pairing them with the positive labels of their original images. When the mismatched feature-label pair of minority attributes are over-sampled by the \emph{Stage\#2} of FRDL, noise can be greatly overfitted by the classifier, to which we refer as the feature noise in FRDL. In Appendix \ref{C1}, we prove following proposition,
 
\begin{proposition}
Eq.\ref{augloss} is upper bounded by the optimum feature de-noising BCE loss:
\begin{equation}
\begin{aligned}
 \frac{1}{N}\sum_{i=1}^N\sum_{k=1}^C |\mathcal L_{cls}(\boldsymbol f^k_i) + \log (1 - \sigma^k) |,
\end{aligned}
\end{equation}
where $\sigma^k$ represents the feature noise rate of attribute $k$.  
\end{proposition}

Consequently, we also apply Eq.\ref{augloss} in the \emph{Stage\#2} of FRDL to train the label-balanced classifier, thereby rendering less overfitting on the spurious features. Finally, GOAT can be realized implicitly to enhance FRDL in term of semantics diversification and feature de-noising, and Algorithm \ref{alg} overviews the whole workflow, where we ignore the classifier bias for brevity, and $W = (\boldsymbol w^1, \boldsymbol w^2, ..., \boldsymbol w^C)$.  

\section{Experiments}
\subsection{Experimental Setup}

\textbf{Evaluation protocol.} We perform experiments on popular large-scale PAR datasets of PA100k \cite{liu2017hydraplus}, PETA \cite{2014Pedestrian} and RAPv1 \cite{li2016richly}. For the datasets configuration, we strictly follow \cite{2020Rethinking} to make a wide and fair comparison with prior arts. It is noteworthy that, for this datasets protocol, there are total 60 annotated attributes in PETA, but 25 attributes are dismissed from evaluation due to their great label asymmetry. For RAPv1, 21 attributes are disregarded for the same reason. Considering that some dropped attributes only have a handful of samples, also to be consistent with the datasets configuration of prior arts, we do not use full attributes of PETA and RAP in our testing as well. We discern between methods by reporting their scores on the label-based metric mean Accuracy (mA), which computes the mean of all attributes recognition accuracy on the positive and negative data. Instance-based metric F1-score (F1) is also evaluated. Details are placed in Appendix \ref{B1} and Appendix \ref{D1}.

\begin{figure}[t]
\centering
\includegraphics[height=4.15cm,width=7.85cm]{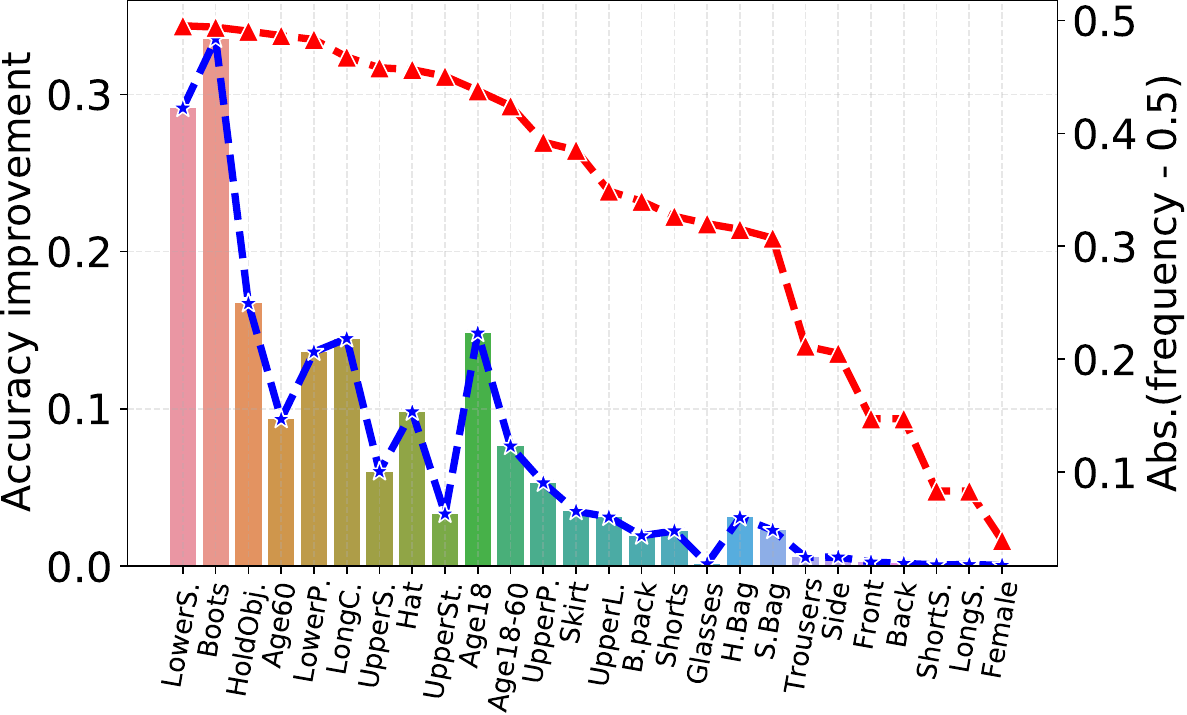}
\caption{The boost of attributes mA of our method w.r.t the baseline for PA100k. Attributes are placed in a decreasing order of the absolute difference between their label mean and 0.5 (red-dashed line), which can quantitatively measure their label imbalance.}
\label{output}
\end{figure}

\textbf{Implementation details.} We adopt ConvNeXt-base \cite{liu2022convnet} as the backbone of $\mathcal H_{\theta}(\cdot)$, due to its desirable trade-off between performance and efficiency. Classifiers and $\mathcal T_{\psi}(\cdot)$ are instantiated by single fully-connected layer as the simplest form among possible variants. Image is spatially resized to 256$\times$192 for input, and batch size is set as 64. Adam solver is applied with weight decay of 5e-4. Horizontal flip and random crop are the only image augmentation methods. The learning rate starts at 1e-4 and decays by a factor of 10 at certain steps.  Unless otherwise stated, we on default set the T in Algorithm \ref{alg} as 20. Other details can be referred in our code at \href{https://github.com/SDret/Pedestrian-Attribute-Recognition-as-Label-balanced-Multi-label-Learning}{github}.

\subsection{Benchmark Results} 

We conduct a thorough evaluation of our method, comparing with strong baselines and a range of recent approaches. The results are presented in Table \ref{benchmark}. Basically, our method enjoys a wide range of meritorious superiorities with practical significance in the challenging real-world scenarios:

\CIRCLE\ \textbf{Strong performance.} FRDL and GOAT in tandem excel existing methods utterly in mA on PA100k and RAPv1. While for other settings, our method is at least on-par with others. Also, prior arts comparable scores on PETA might attribute to the data leakage in its training set about the test data \cite{jia2021rethinking}, and thus are likely overrated \cite{zhou2023solution}. In Appendix \ref{B3}, when the data leakage on PETA is tackled, our method outperforms prior methods with considerable margins. Overall, the result not only highlights the effectiveness of our proposal, but also signifies that modern PARs do not fuel significantly performant models due to their ineffective treatment of the pivotal label imbalances, reinforcing the driving principle of this paper. 

\CIRCLE\ \textbf{Good generalizability.} Our method emphasizes on a general problems of the asymmetry in label distribution, thereby functioning with less inductive biases. Figure \ref{output} reports the attribute-wise accuracy increase of our method over the baseline, and we observe sizable improvements on all attributes: performance gain is larger for infrequent attributes, less for balanced attributes, but never negative. 

\CIRCLE\ \textbf{Minimal computational burden.} Unlike prior arts, we do not facilitate PAR in a multi-modal or multi-task manner by involving related tasks, and nor do we pay a premium regarding parameters by stacking costly modules. During inference, our method exercises with the computational footprint as minimal as that of any baseline model, but still yields overall best accuracy, without the bells and whistles.

\CIRCLE\ \textbf{High compatibility.} Both FRDL and GOAT are macro learning pipelines that lean on no specific or customized network architectures. Thus, our work is of great applicability and can be employed as an effortless plug-and-play companion onto any existing methods.

\begin{table}
	\caption{Comparison of our methods with their existing alternatives, and the ablation results. All values are mA.}
	\label{as}
	\centering
	\scalebox{0.79}{
	\begin{tabular}{ccc|ccc}
	 \toprule
	 \multicolumn{3}{c}{\multirow{1}{*}{Method}}&\multicolumn{1}{c}{PA100k}&\multicolumn{1}{c}{RAPv1}&\multicolumn{1}{c}{PETA}\\	
	 \midrule
	 \multicolumn{3}{c|}{Baseline}&\multicolumn{1}{c}{82.30}&\multicolumn{1}{c}{80.61}&\multicolumn{1}{c}{86.29}\\
	 \hline
	 \multicolumn{3}{c|}{re-weighting\#1 \cite{li2015multi}}&\multicolumn{1}{c}{84.29}&\multicolumn{1}{c}{83.11}&\multicolumn{1}{c}{87.46}\\
	 \multicolumn{3}{c|}{re-weighting\#2 \cite{tan2020relation}}&\multicolumn{1}{c}{84.42}&\multicolumn{1}{c}{83.47}&\multicolumn{1}{c}{87.67}\\
	 \multicolumn{3}{c|}{re-weighting\#3 \cite{zhangdistribution}}&\multicolumn{1}{c}{84.77}&\multicolumn{1}{c}{83.55}&\multicolumn{1}{c}{\textbf{87.85}}\\
	 \multicolumn{3}{c|}{\textbf{FRDL}(Ours)}&\multicolumn{1}{c}{\textbf{88.53}}&\multicolumn{1}{c}{\textbf{86.31}}&\multicolumn{1}{c}{87.55}\\
	\hline
	\multicolumn{3}{c|}{\textbf{FRDL}+ISDA\cite{wang2019implicit}}&\multicolumn{1}{c}{88.70}&\multicolumn{1}{c}{86.52}&\multicolumn{1}{c}{87.94}\\
	\multicolumn{3}{c|}{\textbf{FRDL}+ISDA ($\boldsymbol \Sigma^{*}$ is random noise)}&\multicolumn{1}{c}{88.79}&\multicolumn{1}{c}{86.46}&\multicolumn{1}{c}{87.93}\\
	\multicolumn{3}{c|}{\textbf{FRDL}+MetaSAug\cite{li2021metasaug}}&\multicolumn{1}{c}{88.63}&\multicolumn{1}{c}{86.58}&\multicolumn{1}{c}{87.80}\\
	 \midrule
	 \multicolumn{3}{c|}{\textbf{FRDL}+\textbf{GOAT} of}&\multirow{2}{*}{PA100k}&\multirow{2}{*}{RAPv1}&\multirow{2}{*}{PETA}\\
	 \multicolumn{1}{c}{Eq.\ref{augment}}&\multicolumn{1}{c}{Eq.\ref{augloss} (SGD)}&\multicolumn{1}{c|}{feature de-noising}\\
	 \midrule
	 \multicolumn{1}{c}{\checkmark}&\multicolumn{1}{c}{}&\multicolumn{1}{c|}{}&\multicolumn{1}{c}{88.84}&\multicolumn{1}{c}{86.75}&\multicolumn{1}{c}{87.96}\\
	\multicolumn{1}{c}{}&\multicolumn{1}{c}{\checkmark}&\multicolumn{1}{c|}{}&\multicolumn{1}{c}{89.15}&\multicolumn{1}{c}{87.52}&\multicolumn{1}{c}{88.27}\\
	\multicolumn{1}{c}{}&\multicolumn{1}{c}{\checkmark}&\multicolumn{1}{c|}{\checkmark}&\multicolumn{1}{c}{\textbf{89.44}}&\multicolumn{1}{c}{\textbf{87.72}}&\multicolumn{1}{c}{\textbf{88.59}}\\
	 \bottomrule
	\end{tabular}}
\end{table}

\begin{figure}[t]    
  \centering      
  \subfigure[]
  {
      \label{frdl_upper}\includegraphics[width=0.299\textwidth]{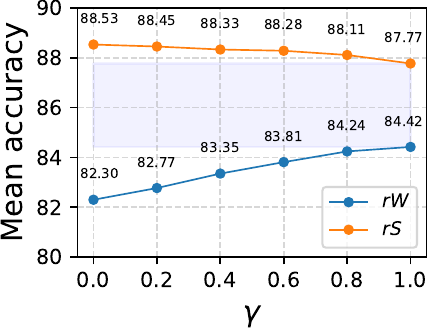}
  }
  \subfigure[]
  {
      \label{ourc}\includegraphics[width=0.14\textwidth]{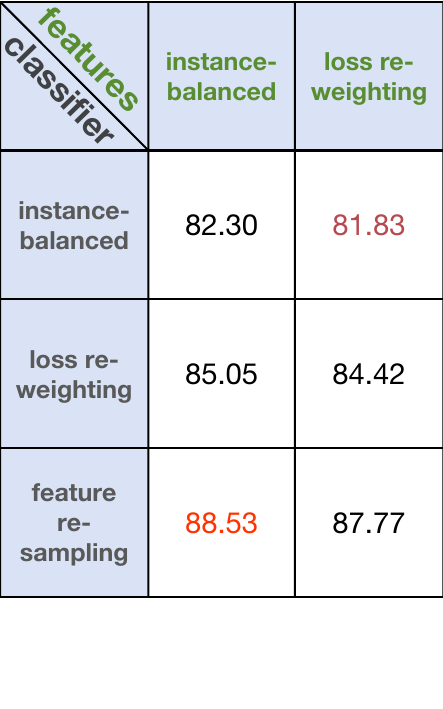}
  }
\caption{(a): We contrast the representation quality of backbones learned with various label-balancing ratio $\gamma$ on PA100k. From $\gamma$ of 0 to 1, feature extractor transitions from that of a plain baseline to that of the loss re-weighted model. The y-axis shows the mA of a classifier re-trained on each of these converged feature extractors, by which we discern between the feature quality of them. $rS$ denotes that the classifier is re-trained by the \emph{Stage\#2} of FRDL, while $rW$ represents the classifier is trained by loss re-weighting. (b): The mA of different feature extractor + classifier combinations on PA100k. For PAR, it manifests that: (1) label balancing obstructs a performant feature extractor; (2) label-balanced feature re-sampling produces fairly better classifier; (3) LIR is supposed to lie between 84.42 - 87.77\% mA, falling behind FRDL (88.53\%).} 
\end{figure}

\begin{figure*}[t]    
  \centering      
  \subfigure[]
  {
      \label{isdac}\includegraphics[width=0.15\textwidth]{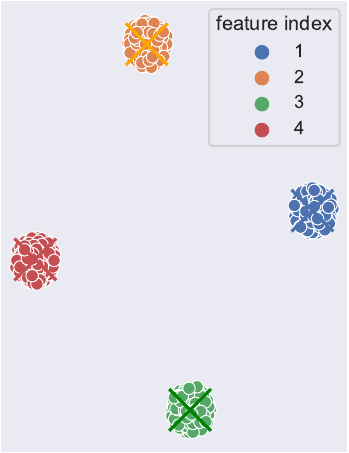}
  }
  \subfigure[]
  {
      \label{ourc}\includegraphics[width=0.15\textwidth]{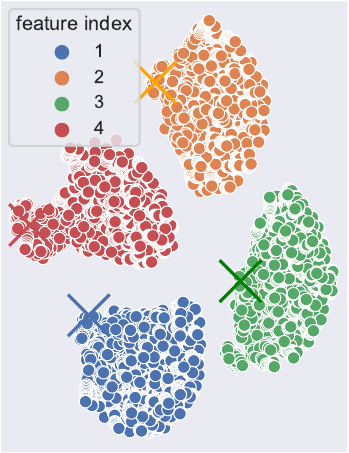}
  }
    \subfigure[]
  {
      \label{eigen}\includegraphics[width=0.29\textwidth]{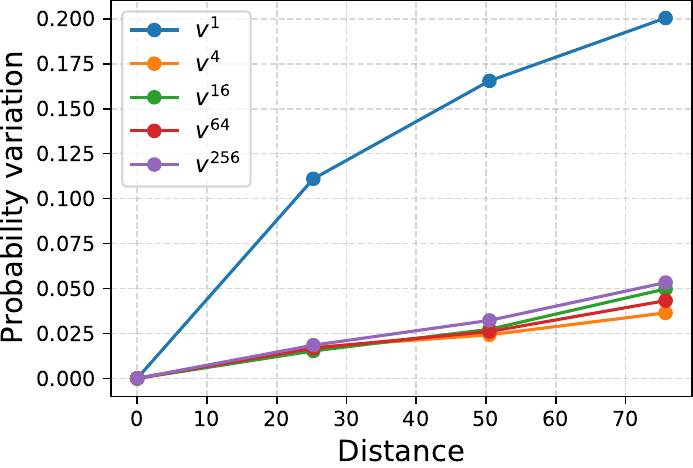}
  }
    \subfigure[]
  {
      \label{dist}\includegraphics[width=0.315\textwidth]{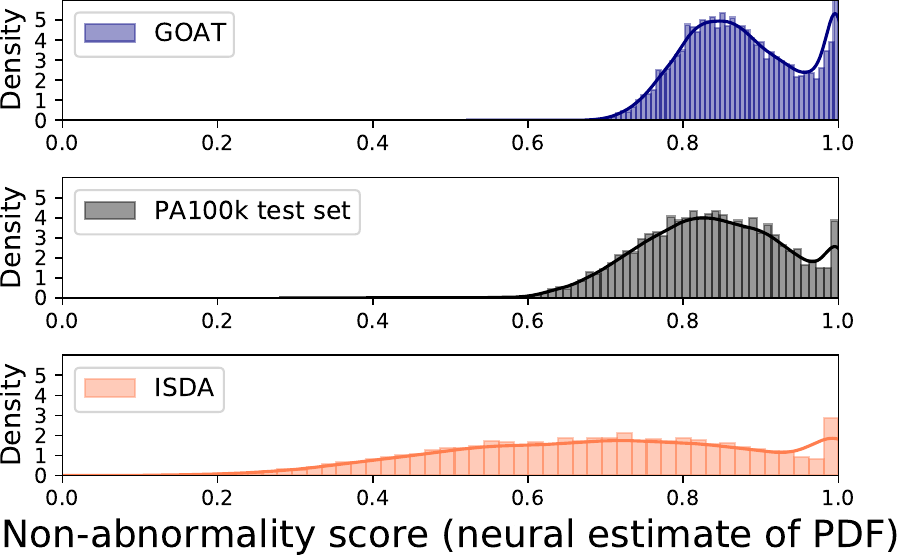}
  }
  \caption{Proof-of-concept experiments for GOAT. T-SNE map of the augmented features distribution for (a) ISDA and (b) GOAT ($\times$ denotes the original feature). (c): Posterior variation along the rays corresponding to different eigenvectors of GOAT translation covariance matrix. (d): For PA100k, the distributions of the PDF scores for the features from test data, ISDA and GOAT augmentations, respectively.}  
  \label{fig:subfig}           
\end{figure*}

\subsection{FRDL Achieves Label Balance}
Present works to dampen the label imbalance in multi-label tasks rely on loss re-weighting techniques. In Table \ref{as}, FRDL competes with some of the best performing re-weighting functions for PAR \cite{jia2021rethinking}, which scale attributes loss by their labels mean, and have been widely integrated into notable works \cite{lu2023orientation, zhou2023solution}. In Table \ref{as}, FRDL outscores both the baseline and re-weightings with substantial margins. Also, it is noteworthy that re-weighting alone brings about 1.5-3\% improvement of mA, which is not trivial as the total improvements of prior methods over baseline vary within about 1-4\% mA. Consequentially, it double-verifies our point of view that label imbalance is the main performance bottleneck for PAR.

Moreover, we state that FRDL develops true label balancing for PAR, since it always delivers performance better than LIR. As is validated by \cite{kang2019decoupling}: (1) learning the feature extractor with instance-balanced sampling produces more generalizable features; (2) learning the classifier with label-balanced sampling sets proper decision boundaries over the learned representations. Correspondingly, if the PAR model is decoupled into a feature extractor of $(\theta, \psi)$ and classifiers denoted by $W$, for FRDL, $(\theta, \psi)$ is trained by plain instance-balanced sampling, while $W$ is updated with label-balanced features, meeting the expectations of both (1) and (2). Hence, by inferencing on a better feature extractor, FRDL is an empirical upper replacement of LIR.

We experimentally prove it in Figure \ref{frdl_upper}, where we train a sequence of feature extractors on PA100k with different label-balancing ratio (detailed in Appendix \ref{D2}), and examine the feature quality of them by comparing the accuracy of classifiers re-trained atop their representations. Since a perfectly label-balanced feature extractor of PAR is practically impossible due to Eq.\ref{condition}, we apply loss re-weighting \cite{tan2020relation} in this study to simulate $(\theta, \psi)$ learned from relatively balanced labels. In Figure \ref{frdl_upper}, by decreasing the degree of label balancing in feature extractor training, the feature quality upgrades persistently, and the best feature is obtained at $(\theta, \psi)$ from instance-balanced sampling, which is exactly FRDL. Since the $(\theta, \psi)$ of LIR would be fully label-balanced trained, its accuracy corresponds to a point in the blue-shaded region of Figure \ref{frdl_upper}, revealing an inferior performance of LIR when competed with FRDL. Thus, as it is intractable and unnecessary to disentangle attributes in image, FRDL could serve as an optimum paradigm for label balancing in multi-label visual tasks.

\subsection{GOAT Approximates Bayesian Feature Sampling} \label{ej}

In Table \ref{as}, ISDA and MetaSAug do not give much boosting over FRDL. It is expectable, since the homogeneity assumption of them is too strong and invites their equivalence to constant-margin optimizer. We experimentally justify it by replacing the feature-sampling covariance $\boldsymbol \Sigma$ in ISDA with random gaussian noise $\boldsymbol \Sigma^{*}$, and fine-tune the $\lambda$ to get best results. It reports that specific form of $\boldsymbol \Sigma$ does not make much difference in the final results, as its values would be anyway balanced out, by the optimized $\lambda^*$, to achieve the expected margin of $\boldsymbol w^{\top} \lambda \boldsymbol \Sigma \boldsymbol w = \boldsymbol w^{\top} \lambda^* \boldsymbol \Sigma^{*} \boldsymbol w$, i.e., what matters for prior arts is not the design of sampling distribution, but the carefully tuned final inter-label margin in Eq.\ref{constant}. 

In contrast, when switching to GOAT, the performance of FRDL is fostered about 1\% mA, indicating that the heterogeneously exploited semantics can milden its overfitting to some extent. For a pictorial grasp of it, we exemplify some sampling distributions of ISDA and GOAT in Figure \ref{isdac}-\ref{ourc}. It shows that the synthetic representations of GOAT are heterogeneously scattered, while those from ISDA simply form repeated gaussian clouds encircling original points. Moreover, GOAT enjoys additional stochasticity from the fact that we run its iterates in a SGD manner. To examine the quality of such stochastic representations, we study the posterior variation along the directions corresponding to the eigenvectors of the heterogeneous cloud. In detail, we use the randomized SVD \cite{halko2011finding} to compute the eigenvectors of the covariance matrix of 1024 translating directions sampled from the heterogeneous cloud of a feature $\boldsymbol f_i$. Then, we calculate the attributes posterior as a function of the distance $t$ from $\boldsymbol f_i$ along its $l$-th eigenvector $\boldsymbol v^l_i$, and visualize the expectation of it under all features from the training set $\mathbb E_{\boldsymbol f_i}[|\mathcal G(\boldsymbol f_i + t\cdot \frac{\boldsymbol v^l_i}{\Vert \boldsymbol v^l_i \Vert }) - \mathcal G(\boldsymbol f_i)|]$ in Figure \ref{eigen}, where $\mathcal G(\cdot)$ represents the classifier function to give sigmoid probabilities. It discovers a strong correlation between the main variance of the GOAT features and the attributes informativeness. In other words, GOAT iterates co-linear with the local geometry of semantics transition, and thus ensure the augmented data semantically novel from its initials. 

Although GOAT produces meaningful features that are distinct from their originals, the augmented points are still in-distribution. This is because that GOAT solely translates feature along its high-density direction, evident in Figure \ref{ourc} that the initial features predominantly reside at the peripheries of their clouds. To confirm it, we apply \cite{2022Rethinking} to estimate the probability density function (PDF) of the features in PA100k training data, and utilize this PDF as an in-distribution metric to quantify the non-abnormality of augmented features. Our findings, presented in the Figure \ref{dist}, demonstrate a significant overlap between the PDFs of GOAT features and the inlier features from PA100k test images, indicating that our method is safe in term of not generating outliers. In this regard, GOAT is endued with the Bayesian feature sampling power to give probabilistic representations. A further discussion is in Appendix \ref{D3}.

\section{Conclusion}
We show that label imbalance is the overlooked \emph{grey rhino} that primarily hinders PAR on realistic datasets. We address this long-standing issue by proposing two complementary methods, FRDL and GOAT, to facilitate unprecedented label balancing and ameliorate the consequential semantics imbalance, in a highly unified framework. Comprehensive discussion and experiments underscore our proposals state-of-the-art outperforming and compelling applicabilities: it is generic, lightweight, simple, catering and orthogonal to previous architectural approaches. At a higher level, label imbalance is a thorny problem for numerous multi-label tasks, endorsing our work shedding light not only on PAR, but a wide array of real-world multi-label recognitions so.  

\section*{Acknowledgements}
This work was partially supported by the "Pioneer" and "Leading Goose" R\&D Program of Zhejiang (Grant No. 2023C01030), the National Natural Science Foundation of China (No.62122011, U21A20514), and the Fundamental Research Funds for the Central Universities. (Corresponding Author: Hai-Miao Hu)


\bibliography{example_paper}

\begin{thebibliography}{55}
\providecommand{\natexlab}[1]{#1}
\providecommand{\url}[1]{\texttt{#1}}
\expandafter\ifx\csname urlstyle\endcsname\relax
  \providecommand{\doi}[1]{doi: #1}\else
  \providecommand{\doi}{doi: \begingroup \urlstyle{rm}\Url}\fi

\bibitem[Bao et~al.(2023)Bao, Wei, Qiu, Zhou, Li, and Tian]{bao2023learning}
Bao, L., Wei, L., Qiu, X., Zhou, W., Li, H., and Tian, Q.
\newblock Learning transferable pedestrian representation from multimodal
  information supervision.
\newblock \emph{arXiv preprint arXiv:2304.05554}, 2023.

\bibitem[Cao et~al.(2023)Cao, Fang, Zhang, Hou, Zhang, and Huang]{cao2023novel}
Cao, Y., Fang, Y., Zhang, Y., Hou, X., Zhang, K., and Huang, W.
\newblock A novel self-boosting dual-branch model for pedestrian attribute
  recognition.
\newblock \emph{Signal Processing: Image Communication}, 115:\penalty0 116961,
  2023.

\bibitem[Chawla et~al.(2002)Chawla, Bowyer, Hall, and
  Kegelmeyer]{chawla2002smote}
Chawla, N.~V., Bowyer, K.~W., Hall, L.~O., and Kegelmeyer, W.~P.
\newblock Smote: synthetic minority over-sampling technique.
\newblock \emph{Journal of artificial intelligence research}, 16:\penalty0
  321--357, 2002.

\bibitem[Chen et~al.(2022)Chen, Jiang, Li, and Wang]{chen2022improving}
Chen, Q., Jiang, W., Li, K., and Wang, Y.
\newblock Improving energy-based out-of-distribution detection by sparsity
  regularization.
\newblock In \emph{Pacific-Asia Conference on Knowledge Discovery and Data
  Mining}, pp.\  539--551. Springer, 2022.

\bibitem[Cormier et~al.(2023)Cormier, Specker, Junior, Jacques, Florin,
  Metzler, Moeslund, Nasrollahi, Escalera, and Beyerer]{cormier2023upar}
Cormier, M., Specker, A., Junior, J., Jacques, C., Florin, L., Metzler, J.,
  Moeslund, T.~B., Nasrollahi, K., Escalera, S., and Beyerer, J.
\newblock Upar challenge: Pedestrian attribute recognition and attribute-based
  person retrieval--dataset, design, and results.
\newblock In \emph{Proceedings of the IEEE/CVF Winter Conference on
  Applications of Computer Vision}, pp.\  166--175, 2023.

\bibitem[Cubuk et~al.(2019)Cubuk, Zoph, Mane, Vasudevan, and
  Le]{2019AutoAugment}
Cubuk, E.~D., Zoph, B., Mane, D., Vasudevan, V., and Le, Q.~V.
\newblock Autoaugment: Learning augmentation strategies from data.
\newblock In \emph{2019 IEEE/CVF Conference on Computer Vision and Pattern
  Recognition (CVPR)}, 2019.

\bibitem[Deng et~al.(2014)Deng, Ping, Chen, and Tang]{2014Pedestrian}
Deng, Y., Ping, L., Chen, C.~L., and Tang, X.
\newblock Pedestrian attribute recognition at far distance.
\newblock \emph{ACM}, 2014.

\bibitem[DeVries \& Taylor(2017{\natexlab{a}})DeVries and
  Taylor]{devries2017dataset}
DeVries, T. and Taylor, G.~W.
\newblock Dataset augmentation in feature space.
\newblock \emph{arXiv preprint arXiv:1702.05538}, 2017{\natexlab{a}}.

\bibitem[DeVries \& Taylor(2017{\natexlab{b}})DeVries and
  Taylor]{devries2017improved}
DeVries, T. and Taylor, G.~W.
\newblock Improved regularization of convolutional neural networks with cutout.
\newblock \emph{arXiv preprint arXiv:1708.04552}, 2017{\natexlab{b}}.

\bibitem[Fabbri et~al.(2017)Fabbri, Calderara, and
  Cucchiara]{fabbri2017generative}
Fabbri, M., Calderara, S., and Cucchiara, R.
\newblock Generative adversarial models for people attribute recognition in
  surveillance.
\newblock In \emph{2017 14th IEEE international conference on advanced video
  and signal based surveillance (AVSS)}, pp.\  1--6. IEEE, 2017.

\bibitem[Fan et~al.(2020)Fan, Hu, Liu, Lu, and Pu]{2020Correlation}
Fan, H., Hu, H.~M., Liu, S., Lu, W., and Pu, S.
\newblock Correlation graph convolutional network for pedestrian attribute
  recognition.
\newblock \emph{IEEE Transactions on Multimedia}, PP\penalty0 (99):\penalty0
  1--1, 2020.

\bibitem[Fan et~al.(2023)Fan, Zhang, Lu, and Wang]{fan2023parformer}
Fan, X., Zhang, Y., Lu, Y., and Wang, H.
\newblock Parformer: Transformer-based multi-task network for pedestrian
  attribute recognition.
\newblock \emph{IEEE Transactions on Circuits and Systems for Video
  Technology}, 2023.

\bibitem[Gal \& Ghahramani(2016)Gal and Ghahramani]{gal2016dropout}
Gal, Y. and Ghahramani, Z.
\newblock Dropout as a bayesian approximation: Representing model uncertainty
  in deep learning.
\newblock In \emph{international conference on machine learning}, pp.\
  1050--1059. PMLR, 2016.

\bibitem[Gal et~al.(2017)Gal, Hron, and Kendall]{gal2017concrete}
Gal, Y., Hron, J., and Kendall, A.
\newblock Concrete dropout.
\newblock \emph{Advances in neural information processing systems}, 30, 2017.

\bibitem[Guo \& Wang(2021)Guo and Wang]{guo2021long}
Guo, H. and Wang, S.
\newblock Long-tailed multi-label visual recognition by collaborative training
  on uniform and re-balanced samplings.
\newblock In \emph{Proceedings of the IEEE/CVF Conference on Computer Vision
  and Pattern Recognition}, pp.\  15089--15098, 2021.

\bibitem[Guo et~al.(2020)Guo, Zheng, Fan, Yu, and Wang]{2020Visual}
Guo, H., Zheng, K., Fan, X., Yu, H., and Wang, S.
\newblock Visual attention consistency under image transforms for multi-label
  image classification.
\newblock In \emph{2019 IEEE/CVF Conference on Computer Vision and Pattern
  Recognition (CVPR)}, 2020.

\bibitem[Halko et~al.(2011)Halko, Martinsson, and Tropp]{halko2011finding}
Halko, N., Martinsson, P.-G., and Tropp, J.~A.
\newblock Finding structure with randomness: Probabilistic algorithms for
  constructing approximate matrix decompositions.
\newblock \emph{SIAM review}, 53\penalty0 (2):\penalty0 217--288, 2011.

\bibitem[Jia et~al.(2021{\natexlab{a}})Jia, Chen, and Huang]{Jia_2021_ICCV}
Jia, Chen, X., and Huang, K.
\newblock Spatial and semantic consistency regularizations for pedestrian
  attribute recognition.
\newblock In \emph{Proceedings of the IEEE/CVF International Conference on
  Computer Vision (ICCV)}, pp.\  962--971, October 2021{\natexlab{a}}.

\bibitem[Jia et~al.(2020)Jia, Huang, Yang, Chen, and Huang]{2020Rethinking}
Jia, J., Huang, H., Yang, W., Chen, X., and Huang, K.
\newblock Rethinking of pedestrian attribute recognition: Realistic datasets
  with efficient method.
\newblock \emph{arXiv}, 2020.

\bibitem[Jia et~al.(2021{\natexlab{b}})Jia, Huang, Chen, and
  Huang]{jia2021rethinking}
Jia, J., Huang, H., Chen, X., and Huang, K.
\newblock Rethinking of pedestrian attribute recognition: A reliable evaluation
  under zero-shot pedestrian identity setting.
\newblock \emph{arXiv preprint arXiv:2107.03576}, 2021{\natexlab{b}}.

\bibitem[Jia et~al.(2022)Jia, Gao, He, Chen, and
  Huang]{DBLP:conf/aaai/JiaGHCH22}
Jia, J., Gao, N., He, F., Chen, X., and Huang, K.
\newblock Learning disentangled attribute representations for robust pedestrian
  attribute recognition.
\newblock pp.\  1069--1077. {AAAI} Press, 2022.

\bibitem[Kang et~al.(2019)Kang, Xie, Rohrbach, Yan, Gordo, Feng, and
  Kalantidis]{kang2019decoupling}
Kang, B., Xie, S., Rohrbach, M., Yan, Z., Gordo, A., Feng, J., and Kalantidis,
  Y.
\newblock Decoupling representation and classifier for long-tailed recognition.
\newblock \emph{arXiv preprint arXiv:1910.09217}, 2019.

\bibitem[Kingma \& Welling(2013)Kingma and Welling]{kingma2013auto}
Kingma, D.~P. and Welling, M.
\newblock Auto-encoding variational bayes.
\newblock \emph{arXiv preprint arXiv:1312.6114}, 2013.

\bibitem[Li et~al.(2015)Li, Chen, and Huang]{li2015multi}
Li, D., Chen, X., and Huang, K.
\newblock Multi-attribute learning for pedestrian attribute recognition in
  surveillance scenarios.
\newblock In \emph{2015 3rd IAPR Asian Conference on Pattern Recognition
  (ACPR)}, pp.\  111--115. IEEE, 2015.

\bibitem[Li et~al.(2016)Li, Zhang, Chen, Ling, and Huang]{li2016richly}
Li, D., Zhang, Z., Chen, X., Ling, H., and Huang, K.
\newblock A richly annotated dataset for pedestrian attribute recognition.
\newblock \emph{arXiv preprint arXiv:1603.07054}, 2016.

\bibitem[Li et~al.(2017)Li, Chen, Zhang, and Huang]{2017Learning}
Li, D., Chen, X., Zhang, Z., and Huang, K.
\newblock Learning deep context-aware features over body and latent parts for
  person re-identification.
\newblock \emph{IEEE}, 2017.

\bibitem[Li et~al.(2021)Li, Gong, Liu, Wang, Qiao, and Cheng]{li2021metasaug}
Li, S., Gong, K., Liu, C.~H., Wang, Y., Qiao, F., and Cheng, X.
\newblock Metasaug: Meta semantic augmentation for long-tailed visual
  recognition.
\newblock In \emph{Proceedings of the IEEE/CVF conference on computer vision
  and pattern recognition}, pp.\  5212--5221, 2021.

\bibitem[Li et~al.(2022)Li, Cao, Feng, Zhou, and Lu]{2022Label2Label}
Li, W., Cao, Z., Feng, J., Zhou, J., and Lu, J.
\newblock Label2label: A language modeling framework for multi-attribute
  learning.
\newblock In \emph{Computer Vision--ECCV 2022: 17th European Conference, Tel
  Aviv, Israel, October 23--27, 2022, Proceedings, Part XII}, pp.\  562--579.
  Springer, 2022.

\bibitem[Liu et~al.(2018)Liu, Liu, Yan, and Shao]{liu2018localization}
Liu, P., Liu, X., Yan, J., and Shao, J.
\newblock Localization guided learning for pedestrian attribute recognition.
\newblock \emph{arXiv preprint arXiv:1808.09102}, 2018.

\bibitem[Liu et~al.(2016)Liu, Wen, Yu, and Yang]{liu2016large}
Liu, W., Wen, Y., Yu, Z., and Yang, M.
\newblock Large-margin softmax loss for convolutional neural networks.
\newblock \emph{arXiv preprint arXiv:1612.02295}, 2016.

\bibitem[Liu et~al.(2017)Liu, Zhao, Tian, Sheng, Shao, Yi, Yan, and
  Wang]{liu2017hydraplus}
Liu, X., Zhao, H., Tian, M., Sheng, L., Shao, J., Yi, S., Yan, J., and Wang, X.
\newblock Hydraplus-net: Attentive deep features for pedestrian analysis.
\newblock In \emph{Proceedings of the IEEE international conference on computer
  vision}, pp.\  350--359, 2017.

\bibitem[Liu et~al.(2022)Liu, Mao, Wu, Feichtenhofer, Darrell, and
  Xie]{liu2022convnet}
Liu, Z., Mao, H., Wu, C.-Y., Feichtenhofer, C., Darrell, T., and Xie, S.
\newblock A convnet for the 2020s.
\newblock In \emph{Proceedings of the IEEE/CVF Conference on Computer Vision
  and Pattern Recognition}, pp.\  11976--11986, 2022.

\bibitem[Lu et~al.(2023)Lu, Hu, Yu, Zhou, Wang, and Li]{lu2023orientation}
Lu, W.-Q., Hu, H.-M., Yu, J., Zhou, Y., Wang, H., and Li, B.
\newblock Orientation-aware pedestrian attribute recognition based on graph
  convolution network.
\newblock \emph{IEEE Transactions on Multimedia}, 2023.

\bibitem[Nguyen et~al.(2022)Nguyen, Lu, Munoz, Raff, Nicholas, and
  Holt]{nguyen2022out}
Nguyen, A.~T., Lu, F., Munoz, G.~L., Raff, E., Nicholas, C., and Holt, J.
\newblock Out of distribution data detection using dropout bayesian neural
  networks.
\newblock In \emph{Proceedings of the AAAI Conference on Artificial
  Intelligence}, volume~36, pp.\  7877--7885, 2022.

\bibitem[Specker et~al.(2022)Specker, Cormier, and Beyerer]{Specker2022UPARUP}
Specker, A., Cormier, M., and Beyerer, J.
\newblock Upar: Unified pedestrian attribute recognition and person retrieval.
\newblock \emph{ArXiv}, abs/2209.02522, 2022.

\bibitem[Tan et~al.(2020{\natexlab{a}})Tan, Yang, Wan, Guo, and
  Li]{tan2020relation}
Tan, Yang, Y., Wan, J., Guo, G., and Li, S.~Z.
\newblock Relation-aware pedestrian attribute recognition with graph
  convolutional networks.
\newblock In \emph{Proceedings of the AAAI conference on artificial
  intelligence}, volume~34, pp.\  12055--12062, 2020{\natexlab{a}}.

\bibitem[Tan et~al.(2020{\natexlab{b}})Tan, Yang, Wan, Guo, and
  Li]{2020Relation}
Tan, Z., Yang, Y., Wan, J., Guo, G., and Li, S.~Z.
\newblock Relation-aware pedestrian attribute recognition with graph
  convolutional networks.
\newblock \emph{Proceedings of the AAAI Conference on Artificial Intelligence},
  34\penalty0 (7):\penalty0 12055--12062, 2020{\natexlab{b}}.

\bibitem[Tang \& Huang(2022)Tang and Huang]{tang2022drformer}
Tang, Z. and Huang, J.
\newblock Drformer: Learning dual relations using transformer for pedestrian
  attribute recognition.
\newblock \emph{Neurocomputing}, 497:\penalty0 159--169, 2022.

\bibitem[Wan et~al.(2018)Wan, Zhong, Li, and Chen]{wan2018rethinking}
Wan, W., Zhong, Y., Li, T., and Chen, J.
\newblock Rethinking feature distribution for loss functions in image
  classification.
\newblock In \emph{Proceedings of the IEEE conference on computer vision and
  pattern recognition}, pp.\  9117--9126, 2018.

\bibitem[Wang et~al.(2017)Wang, Zhu, and Gong]{wang2017discovering}
Wang, J., Zhu, X., and Gong, S.
\newblock Discovering visual concept structure with sparse and incomplete tags.
\newblock \emph{Artificial Intelligence}, 250:\penalty0 16--36, 2017.

\bibitem[Wang et~al.(2022)Wang, Zheng, Yang, Zheng, Chen, Tang, and
  Luo]{wang2022pedestrian}
Wang, X., Zheng, S., Yang, R., Zheng, A., Chen, Z., Tang, J., and Luo, B.
\newblock Pedestrian attribute recognition: A survey.
\newblock \emph{Pattern Recognition}, 121:\penalty0 108220, 2022.

\bibitem[Wang et~al.(2019)Wang, Pan, Song, Zhang, Huang, and
  Wu]{wang2019implicit}
Wang, Y., Pan, X., Song, S., Zhang, H., Huang, G., and Wu, C.
\newblock Implicit semantic data augmentation for deep networks.
\newblock \emph{Advances in Neural Information Processing Systems}, 32, 2019.

\bibitem[Weng et~al.(2023)Weng, Tan, Fang, and Guo]{weng2023exploring}
Weng, D., Tan, Z., Fang, L., and Guo, G.
\newblock Exploring attribute localization and correlation for pedestrian
  attribute recognition.
\newblock \emph{Neurocomputing}, 531:\penalty0 140--150, 2023.

\bibitem[Wu et~al.(2022)Wu, Huang, Gao, Hong, Zhao, and Du]{wu2022inter}
Wu, J., Huang, Y., Gao, Z., Hong, Y., Zhao, J., and Du, X.
\newblock Inter-attribute awareness for pedestrian attribute recognition.
\newblock \emph{Pattern Recognition}, 131:\penalty0 108865, 2022.

\bibitem[Xu et~al.(2022)Xu, Zheng, Zhang, Sun, Li, and Zhu]{xu2022adaptive}
Xu, C., Zheng, Y., Zhang, Y., Sun, C., Li, G., and Zhu, Z.
\newblock Adaptive class-balanced loss based on re-weighting.
\newblock In \emph{2022 6th Asian Conference on Artificial Intelligence
  Technology (ACAIT)}, pp.\  1--8. IEEE, 2022.

\bibitem[Zhang et~al.(2021{\natexlab{a}})Zhang, Bengio, Hardt, Recht, and
  Vinyals]{zhang2021understanding}
Zhang, C., Bengio, S., Hardt, M., Recht, B., and Vinyals, O.
\newblock Understanding deep learning (still) requires rethinking
  generalization.
\newblock \emph{Communications of the ACM}, 64\penalty0 (3):\penalty0 107--115,
  2021{\natexlab{a}}.

\bibitem[Zhang et~al.(2021{\natexlab{b}})Zhang, Han, Cheng, and
  Yang]{zhang2021weakly}
Zhang, D., Han, J., Cheng, G., and Yang, M.-H.
\newblock Weakly supervised object localization and detection: A survey.
\newblock \emph{IEEE transactions on pattern analysis and machine
  intelligence}, 44\penalty0 (9):\penalty0 5866--5885, 2021{\natexlab{b}}.

\bibitem[Zhang et~al.(2017)Zhang, Cisse, Dauphin, and
  Lopez-Paz]{zhang2017mixup}
Zhang, H., Cisse, M., Dauphin, Y.~N., and Lopez-Paz, D.
\newblock mixup: Beyond empirical risk minimization.
\newblock \emph{arXiv preprint arXiv:1710.09412}, 2017.

\bibitem[Zhang et~al.(2021{\natexlab{c}})Zhang, Li, Yan, He, and
  Sun]{zhangdistribution}
Zhang, S., Li, Z., Yan, S., He, X., and Sun, J.
\newblock Distribution alignment: A unified framework for long-tail visual
  recognition (supplementary material).
\newblock 2021{\natexlab{c}}.

\bibitem[Zhang et~al.(2021{\natexlab{d}})Zhang, Wei, Zhou, and
  Wu]{zhang2021bag}
Zhang, Y., Wei, X.-S., Zhou, B., and Wu, J.
\newblock Bag of tricks for long-tailed visual recognition with deep
  convolutional neural networks.
\newblock In \emph{Proceedings of the AAAI conference on artificial
  intelligence}, volume~35, pp.\  3447--3455, 2021{\natexlab{d}}.

\bibitem[Zhang et~al.(2023)Zhang, Kang, Hooi, Yan, and Feng]{zhang2023deep}
Zhang, Y., Kang, B., Hooi, B., Yan, S., and Feng, J.
\newblock Deep long-tailed learning: A survey.
\newblock \emph{IEEE Transactions on Pattern Analysis and Machine
  Intelligence}, 2023.

\bibitem[Zheng et~al.(2023)Zheng, Wang, Wang, Huang, He, and
  Hussain]{zheng2023diverse}
Zheng, A., Wang, H., Wang, J., Huang, H., He, R., and Hussain, A.
\newblock Diverse features discovery transformer for pedestrian attribute
  recognition.
\newblock \emph{Engineering Applications of Artificial Intelligence},
  119:\penalty0 105708, 2023.

\bibitem[Zheng et~al.(2015)Zheng, Shen, Tian, Wang, Bu, and
  Tian]{zheng2015person}
Zheng, L., Shen, L., Tian, L., Wang, S., Bu, J., and Tian, Q.
\newblock Person re-identification meets image search.
\newblock \emph{arXiv preprint arXiv:1502.02171}, 2015.

\bibitem[Zhou(2022)]{2022Rethinking}
Zhou, Y.
\newblock Rethinking reconstruction autoencoder-based out-of-distribution
  detection.
\newblock \emph{Proceedings of the IEEE conference on computer vision and
  pattern recognition}, 2022.

\bibitem[Zhou et~al.(2023)Zhou, Hu, Yu, Xu, Lu, and Cao]{zhou2023solution}
Zhou, Y., Hu, H.-M., Yu, J., Xu, Z., Lu, W., and Cao, Y.
\newblock A solution to co-occurrence bias: Attributes disentanglement via
  mutual information minimization for pedestrian attribute recognition.
\newblock \emph{arXiv preprint arXiv:2307.15252}, 2023.

\end{thebibliography}
\bibliographystyle{icml2024}

\newpage
\appendix
\onecolumn

\section{Overview of Appendix}
The appendix is organized via the following contributions:

Appendix \ref{B} (PAR Datasets) details the adopted datasets and explores our method on additional realistic PAR datasets.
\begin{itemize} 
\item \ref{B1} introduces important statistics about PETA, RAP, PA100k and \textbf{our results on an open-set PAR challenge}.
\item \ref{B2} discusses the data leakage of overlapped pedestrian identities between PETA training set and test set.
\item \ref{B3} \textbf{reports our performance on some realistic datasets}, where the data leakage is well-addressed.
\end{itemize}
Appendix \ref{C} (Theoretical Analysis) elucidates the mathematical insights behind GOAT.
\begin{itemize} 
\item \ref{C1} \textbf{researches the regularization effect of high-density oriented feature translating regarding feature noise}.
\end{itemize}
Appendix \ref{D} (Further Experiments) examines our method with further experimental results.
\begin{itemize} 
\item \ref{D1} explains our performance divergence between mA and F1.
\item \ref{D2} supplements the experimental settings that underlie our results in Figure \ref{frdl_upper} and Figure \ref{fig:subfig}.
\item \ref{D3} \textbf{provides additional insight regarding feature augmentation from the Bayesian point of view}. 
\end{itemize} 

\newpage
\twocolumn

\section{PAR Datasets} \label{B}
\subsection{Basic of PETA, RAP and PA100k}  \label{B1}
PETA, RAP, and PA100k have emerged as three most preeminent datasets for PAR, and are widely adopted by leading methodologies in this domain \cite{DBLP:conf/aaai/JiaGHCH22,tang2022drformer,wu2022inter,weng2023exploring,zheng2023diverse,cao2023novel,bao2023learning,fan2023parformer,lu2023orientation,zhou2023solution}. In Table \ref{dataset}, we present the statistics of these datasets.

\newcolumntype{P}[1]{>{\centering\arraybackslash}p{#1}}
\newcolumntype{M}[1]{>{\centering\arraybackslash}m{#1}}
\begin{table}[h]
\center{}
\footnotesize{
\begin{tabular}{M{2.0cm}M{1.4cm}M{1.4cm}M{1.8cm}}
 \toprule
 & \textbf{PETA} & \textbf{RAP} & \textbf{PA100k} \\
 \midrule
 \midrule
\# sample & 19,000 & 41,585 & 100,000 \\
\# attribute & 35 (60) & 51 (72) & 26 \\
\# scene & - & 26 (indoor) & 598 (outdoor) \\
\# tracklet & - & -  & 18,206 \\
data leakage & \ding{51} & \ding{51} & \ding{56} \\
resolution & $17\times39$ to $169\times365$ & $36\times92$ to $344\times554$ & $50\times100$ to $758\times454$ \\
 \bottomrule
\end{tabular}
}
\caption{Details of the three adopted PAR datasets. The number of attributes denoted within the parentheses signifies the total number of attributes, whereas the numeral presented outside the parentheses represents the experimentally adopted attributes in popular benchmarks. Data leakage indicates whether there is an overlap of pedestrian identities between the training and test set.}
\label{dataset}
\end{table}

\textbf{PETA}. PETA (PEdesTrian Attribute) is introduced by \cite{2014Pedestrian} as a comprehensive dataset encompassing 19,000 meticulously selected images. These images, sourced from ten publicly accessible small-scale datasets, are annotated with 61 binary attributes and four multi-class attributes. Due to the uneven distribution of certain attributes, only 35 of PETA attributes have been kept for evaluation purposes in popular benchmarks.

\textbf{RAP}. \cite{li2016richly} constructed RAP (Richly Annotated Pedestrian) dataset, specifically RAPv1, which comprises 41,585 pedestrian samples. These samples were captured from a real-world surveillance network consisting of 26 video cameras strategically positioned at a busy shopping mall. RAPv1 dataset features detailed annotations for 69 fine-grained attributes, along with annotations for three critical environmental factors: viewpoints, occlusion styles, and body parts. However, for evaluation purposes, only 51 attributes for RAP are chosen in popular benchmarks, based on their proportion of positive samples present.

\textbf{PA100k}. Further advancing the field, PA100k \cite{liu2017hydraplus} is presented with staggering 100,000 images annotated with 26 attributes. PA100k is one of the most extensive pedestrian attribute dataset to date, making it an invaluable resource for a wide range of pedestrian analysis tasks.

\textbf{UPAR}. The UPAR \cite{cormier2023upar} dataset includes 40 crucial binary attributes spanning across 12 distinct attribute categories, and was integrated from four distinct datasets of PA100k, PETA, RAP and Market-1501 \cite{zheng2015person}. UPAR establishes an open-set benchmark for PAR, by training the models on a restricted set of data from specific datasets and subsequently evaluating their performance using data of previously unseen dataset, respecting the realistic deployment environment of PAR models. As the UPAR test set is not released, we use its training set to re-configure a new dataset \textbf{UPAR*}: the UPAR re-labeled PA100k, Market-1501 and PETA are employed as training set, while the re-labeled RAP dataset is leaved out as test set. We report our method on UPAR* in Table \ref{realistic}. 

\begin{table}[h]
	\caption{mA comparison of our method vs. some notable works of VAC \cite{2020Visual}, JLAC \cite{2020Relation}, L2L \cite{2022Label2Label}, OAGCN \cite{lu2023orientation} and PARFormer-B \cite{fan2023parformer} on three realistic datasets of UPAR*, PETAzs and RAPzs. We refer the scores of prior arts on PETAzs and RAPzs from \cite{jia2021rethinking} and \cite{zhou2023solution}. The results of PARFormer-B are produced by the public code from its original literature. We also denote as the subscript of our score the relative improvement over the highest existing method.}
	\label{realistic}
	\centering
	\scalebox{0.75}{
	\begin{tabular}{c|cccccc}
	 \toprule
	 \multicolumn{1}{c|}{Dataset}&VAC&JLAC&L2L&OAGCN&PARFormer&\textbf{Ours}\\
	 \midrule
	 \midrule
	 \rowcolor{lightgray!30} \multicolumn{1}{c|}{UPAR*}&69.85&-&70.44&-&72.58&\textbf{79.40$_{+6.82}$}\\
	 \midrule
	 \multicolumn{1}{c|}{PETA}&84.58&86.88&87.07&88.21&\textbf{88.65}&88.59$_{-0.06}$\\
	  \rowcolor{lightgray!30} \multicolumn{1}{c|}{PETAzs}&71.91&73.60&72.13&75.44&76.16&\textbf{79.10$_{+2.94}$}\\
	  \midrule
	 \multicolumn{1}{c|}{RAP}&80.27&81.51&81.93&86.02&83.84&\textbf{87.72$_{+1.18}$}\\
	 \rowcolor{lightgray!30}  \multicolumn{1}{c|}{RAPzs}&73.70&76.38&73.84&76.20&77.24&\textbf{83.69$_{+6.45}$}\\
	 \bottomrule
	\end{tabular}}
\end{table}

\subsection{Data Leakage in PETA}  \label{B2}
A notable constraint in PETA dataset pertains to the partitioning of training and test sets \cite{jia2021rethinking}. Specifically, PETA images are randomly allocated to either set without any regard for pedestrian identity. Consequently, this haphazard approach in both image acquisition and dataset division results in a great overlap of highly similar images between the training and test sets, with only minor variations in background and pose. This phenomenon, commonly known as "data leakage," poses a challenge in accurately assessing model generalization capabilities, and renders evaluated methods significantly over-estimated.

Similar issue also exists in RAP, however, with a relatively less proportion. While for PA100k dataset, it addressed this issue by assigning all images of a single pedestrian exclusively to either the training or test set. 

\subsection{Realistic Results on PETAzs and RAPzs}  \label{B3}
For reliable performance evaluations of PAR models on PETA and RAP dataset, \cite{jia2021rethinking} undertook re-organization of them and created zero-shot variants dubbed PETAzs and RAPzs. PETAzs and RAPzs adhere strictly to the zero-shot paradigm for pedestrian identities, ensuring no overlap between training and test sets in terms of identities. Subsequently, we have documented our methods mA on these re-configured realistic datasets in Table \ref{realistic}. Our method demonstrates significant superiority over previous works when the issues of data leakage in RAP and PETA are addressed, exhibiting notable margins of improvement.

\section{Theoretical Analysis} \label{C}

\subsection{GOAT Regularizes the Feature Noise of FRDL}  \label{C1}

For brevity, we confine following discussion to the recognition of a single attribute, and the conclusion drawn herein can be readily extrapolated to more conventional instances pertaining to multi-hot labeling. Formally, the \emph{Stage\#2} of FRDL re-samples features on a dataset $\{(\boldsymbol f_i, z_i)\}_{i=1}^N$, where $\boldsymbol f_i$ represents a cached feature to be sampled for the classifier fine-tuning, and $z_i$ is its ground true binary label of a certain attribute. Practically, as $z_i$ is unknown, we use the label $y_i$ that corresponds to its input image $\boldsymbol x_i$ as a proxy, and apply a surrogate dataset $\{(\boldsymbol f_i, y_i)\}_{i=1}^N$ for the implemental feature re-sampling in FRDL. However, since features are extracted with a failure rate $\sigma$, there could be a faulty label assignment of $y_i \neq z_i$ to $\boldsymbol f_i$. Finally, the classifier is re-trained on a polluted dataset, and tends to exhibit poor generalization owing to being misled by the spurious samples that diverge from the true joint distribution of $\{(\boldsymbol f_i, z_i)\}_{i=1}^N$. Assuming that the feature noise ratio $\sigma$ is only label-dependent, above process indicates a conditional probability distribution $P(Y | Z)$ with $P (Y = z | Z = z) = 1 - \sigma$ and $P (Y \neq z | Z = z) = \sigma$. Thus, if we minimize the BCE loss of PAR to re-train a classifier denoted by $W$, we have \vspace{-2.4ex} 

 \begin{small}
 \begin{equation}
\begin{aligned}
&-\frac{1}{N}\sum_{i=1}^{N}\log P(Y = y_i | \boldsymbol f_i; W)\\ 
= &-\frac{1}{N}\sum_{i=1}^{N}\log\sum_{z_i \in \{y_i, \neq y_i\}}P(Y = y_i, Z = z_i  | \boldsymbol f_i; W)\\
= &-\frac{1}{N}\sum_{i=1}^{N}\log\sum_{z_i \in \{y_i,\neq y_i\}}P(Y = y_i  | Z = z_i) P(Z = z_i  | \boldsymbol f_i; W)\\
= &-\frac{1}{N}\sum_{i=1}^{N}\log ((1-\sigma) P(Z = y_i  | \boldsymbol f_i; W) + \sigma P(Z \neq y_i  | \boldsymbol f_i; W))\\
= &-\frac{1}{N}\sum_{i=1}^{N}\log ((1-\sigma) P(Z = y_i  | \boldsymbol f_i; W)\\ 
&\,\,\,\,\,\,\,\,\,\,\,\,\,\,\,\,\,\,\,\,\,\,\,\,\,\,\,\,\,\,\,\,\,\,\,\,\,\,\,\,\,\,\,\,\,\,\,\,\,\,\,\,\,\,\,\,\,\,\,\,\,\,+ \sigma  (1 - P(Z = y_i  | \boldsymbol f_i; W))).\\
\label{noise}
\end{aligned}
\end{equation}
\end{small} \vspace{-2.5ex} 

By taking the derivatives of Eq.\ref{noise}, the BCE loss is minimized when $P(Z = y_i  | \boldsymbol f_i; W) = 1 - \sigma$. It implies that, to relieve the classifier from further overfitting on the noisy features, the following training objective should be minimized w.r.t. $W$ to encourage $P(Z = y_i  | \boldsymbol f_i; W)$ to take $1-\sigma$,   \vspace{-3.4ex}

 \begin{equation}
\begin{aligned}
&\,\,\,\,\,\,\,\,\,\frac{1}{N}\sum_{i=1}^N |-\log P(Z = y_i  | \boldsymbol f_i; W) -  (-\log(1 - \sigma))|\\
&= \frac{1}{N}\sum_{i=1}^N |-\log P(Z = y_i  | \boldsymbol f_i; W)\\
&\,\,\,\,\,\,\,\,\,\,\,\,\,\,\,\,\,\,\,\,\,\,\,\,\,\,\,\,\,\,\,\,\,\,\,\,\,\,\,\,\, - (-\log \mathbb E_{\boldsymbol f} [P(Z = y  | \boldsymbol f; W^*)])|\\
& \ge \frac{1}{N}\sum_{i=1}^N |-\log P(Z = y_i  | \boldsymbol f_i; W) \\
&\,\,\,\,\,\,\,\,\,\,\,\,\,\,\,\,\,\,\,\,\,\,\,\,\,\,\,\,\,\,\,\,\,\,\,\,\,\,\,\,\, - \mathbb E_{\boldsymbol f} [ -\log P(Z = y  | \boldsymbol f; W^*)]|.
\end{aligned}
\label{sur}
\end{equation} \vspace{-2.5ex} 

As $1-\sigma$ denotes the success rate of a feature extractor, it could be estimated as the highest  $\mathbb E_{\boldsymbol f}[P(Z = y  | \boldsymbol f)]$ achievable by a classifier (denoted by $W^*$) on the corresponding features. Considering that neural models lean towards prioritizing fitting clean data before noisy ones \cite{zhang2021understanding}, we simply apply the optimized classifier trained with early stopping in the \emph{Stage\#1} of FRDL as an approximate of $W^*$. After applying Jensen inequality, the optimum feature de-noising objective is exactly an upper bound of our high-density-translating loss in GOAT, regardless that the latter is represented for multi-hot labeling in Eq.\ref{augloss}.

\section{Further Experiments} \label{D}

\subsection{Inconsistency between mA and F1} \label{D1}

Table \ref{benchmark} highlights that our method does not produce F1 scores on par with those of mA. This discrepancy (or inconsistency) between the two evaluation metrics is not unique to our approach. As can be seen in Table \ref{benchmark}, many methodologies that excel in mA also tend to exhibit lower F1 scores. A similar trend has also been summarized in the UPAR challenge \cite{cormier2023upar}, where PAR methods prevailing in terms of mA often falter when assessed using F1. The underlying rationale behind this phenomenon is that mA assigns equal importance to both positive and negative samples when evaluating an attribute, whereas F1 primarily emphasizes the recognition precision of positive labels since it is an instance-based metric:

\begin{small}
\begin{equation}
\begin{aligned}
&Prec = \frac{1}{N}\sum_{i=1}^N \frac{TP_i}{TP_i + FP_i},\\
R&ecall = \frac{1}{N}\sum_{i=1}^N \frac{TP_i}{TP_i + FN_i},\\
&F1 = \frac{2 \cdot Prec \cdot Recall}{Prec + Recall},\\
mA =  \frac{1}{C}&\sum_{j=1}^C \frac{1}{2} (\frac{TP^j}{TP^j + FN^j} + \frac{TN^j}{TN^j + FP^j}), \nonumber
\end{aligned}
\end{equation}
\end{small}

where $TP_i$, $FP_i$, $FN_i$ are the total number of true positive, false
positive and false negative attributes of $i$-th sample, and $TP^j$, $TN^j$, $FP^j$, $FN^j$ are the number of true positive, true negative, false positive and false negative samples of $j$-th attribute. Since every attribute should be regarded as equally important (so as its positive and negative label), mA is typically deemed to have greater practical implication.

\subsection{Experimental Details in Ablations} \label{D2}

In Figure \ref{frdl_upper}, we gradually modified the instance-balanced feature extractor towards that trained with label-balancing technique, and by a comprehensive analysis of the variation of feature quality during this transition, we have concluded that the FRDL method serves as a superior alternative to the LIR approach. Specifically, we applied following weighted BCE loss in the training of all backbones:

\begin{equation}
\begin{aligned}
\mathcal L &= - \sum_{i=1}^N\sum_{j=1}^{C} w_j^i (\boldsymbol y_i^j\log \boldsymbol p_i^j + (1 - \boldsymbol y_i^j)\log(1 - \boldsymbol p_i^j)),\\
&s.t. \,\,\, w_j^i = \left \{
\begin{array}{ll}
    e^{1 -  (\gamma (r_j - 0.5) + 0.5)},                    & \boldsymbol y_i^j = 1\\
    e^{\gamma (r_j - 0.5) + 0.5},                    & \boldsymbol y_i^j = 0
\end{array}
\right.
\nonumber
\end{aligned}
\end{equation}

\begin{figure}[t]
\centering
\includegraphics[height=3.54cm,width=8.2cm]{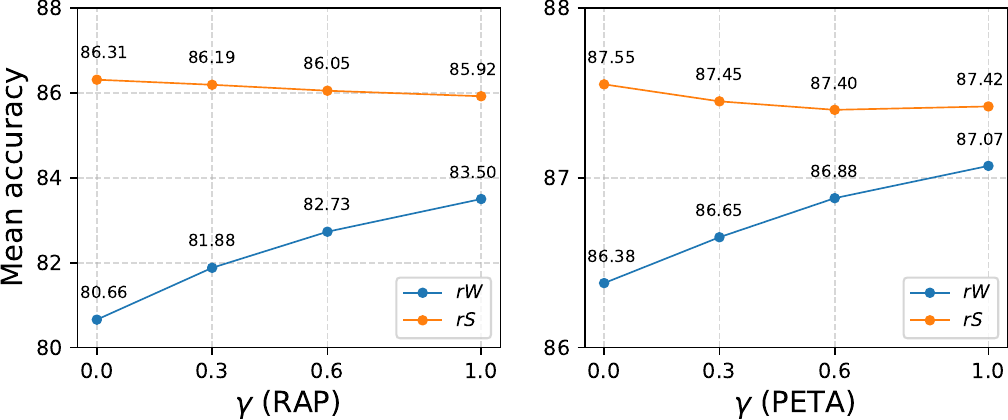}
\caption{Representation quality of feature extractor as a function of the label-balancing ratio $\gamma$ for RAP and PETA, respectively. Implementation details are identical to those for Figure \ref{frdl_upper}.}
\label{all_data_frdl}
\end{figure}

where $ \boldsymbol p_i$ is the estimated attributes posterior of $\boldsymbol x_i$, $r_j$ the label mean of attribute $j$, and $\gamma$ the label balancing ratio that transitions from 0 to 1 to study the impact of label-balancing having on feature extractors. When $\gamma$ is 0, it respects the instance-balanced learning as no label balancing would be exerted. In Figure \ref{all_data_frdl}, we present further experimental results on PETA and RAP datasets, and they exhibit similar trend of variation as that of PA100k in Figure \ref{frdl_upper}.

For the features PDF in Figure \ref{fig:subfig}, we followed \cite{2022Rethinking} to train a feature reconstructor on all activation vectors extracted from PA100k training set. Next, we fit two Weibull distributions on the tail of training-set features reconstruction residual and sigmoid confidence scores, respectively. Finally, we use the Weibulls product as the final feature normality measure. Details are identical to \cite{2022Rethinking, chen2022improving}.

\subsection{Bayesian Inference as Feature Augmentation}  \label{D3}

\begin{table}[h]
	\caption{We compare GOAT with popular data augmentation approaches of AutoAug \cite{2019AutoAugment} (ImageNet policies), Mixup \cite{zhang2017mixup} and Cutout \cite{devries2017improved}, to research its break-down effect as an independent data augmentation method for PAR. GOAT$^{+}$ represents GOAT enhanced by the dropout variational inference.}
	\label{data_aug}
	\centering
	\scalebox{0.75}{
	\begin{tabular}{c|cccccc}
	 \toprule
	 \multicolumn{1}{c|}{Dataset}&Baseline&AutoAug&Mixup&Cutout&GOAT&GOAT$^{+}$\\
	 \midrule
	 \midrule
	 \multicolumn{1}{c|}{PA100k}&82.45&82.55&82.07&81.09&84.16&\textbf{84.50}\\
	 \midrule
	 \multicolumn{1}{c|}{PETAzs}&75.01&75.22&74.50&73.83&76.41&\textbf{76.66}\\
	  \midrule
	 \multicolumn{1}{c|}{RAPzs}&76.14&76.43&75.79&73.56&77.98&\textbf{78.41}\\
	 \bottomrule
	\end{tabular}}
\end{table}

Bayesian approaches quantify uncertainty by assigning a probability distribution to model parameters, and are prevalent in generative modeling frameworks that rely on variational inference \cite{kingma2013auto}, or out-of-distribution detection and uncertainty estimation \cite{nguyen2022out}. Nevertheless, to our knowledge, this study represents a pioneering effort in utilizing the probabilistic characteristic of Bayesian inference to offer additional in-distribution variation for feature augmentation. In Table \ref{data_aug}, we juxtapose GOAT against several prevalent data augmentation strategies. The findings underscore that image augmentation techniques, such as Mixup and Cutout, can potentially corrupt the fine-grained signatures of attributes in the pixel domain, thereby leading to decreased performance in PAR. Conversely, GOAT prevails by manipulating data within the latent space, affirming the indispensability of Bayesian feature augmentation in PAR.

Equipped with Bayesian perspective, the GOAT framework can be further extended by incorporating other Bayesian methodologies. In Table \ref{data_aug}, we demonstrate the application of dropout variational inference \cite{gal2016dropout, gal2017concrete}, which utilizes a spike and slab variational distribution to interpret dropout during the testing phase as an approximation of Bayesian inference, to offer further randomness over the process of feature augmentation.

\end{document}